\documentclass[11pt]{article}
\usepackage{lipsum}

\usepackage[utf8]{inputenc} 
\usepackage[T1]{fontenc}    
\usepackage{hyperref}       
\usepackage{url}            
\usepackage{booktabs}       
\usepackage{amsfonts}       
\usepackage{nicefrac}       
\usepackage{microtype}      
\usepackage{xcolor}         
\usepackage{tikz,times}

\usepackage{framed}
\usepackage{amsthm,nccmath}
\usepackage{amsmath,bm}
\usepackage[normalem]{ulem}
\usepackage{bigints}
\usepackage{amssymb}
\usepackage{caption}
\usepackage{subcaption}
\usepackage{graphicx}
\usepackage{enumitem}
\usepackage{multirow}
\usepackage{color, colortbl}
\usepackage{graphicx}
\usepackage{amssymb}
\usepackage{flushend}
\usepackage[none]{hyphenat}
\usepackage{float}
\usepackage{epstopdf}
\usepackage{footnote}
\makesavenoteenv{tabular}
\makesavenoteenv{table}
\usepackage{bbm}
\usepackage{amsthm}
\usepackage[makeroom]{cancel}
\usepackage{stmaryrd}
\usepackage{mathtools, cuted}
\usepackage{hyperref}

\usepackage[english]{babel}
\usepackage{xspace}
\usepackage{algorithm}
\usepackage[noend]{algorithmic}
\usepackage{tcolorbox}
\usepackage{setspace}

  \providecommand{\tt}{\mathbf{t}}

  \providecommand{\xx}{\mathbf{x}}
  
  \providecommand{\zz}{\mathbf{z}}

  \providecommand{\cD}{\mathcal{D}}

  \providecommand{\cM}{\mathcal{M}}
  \providecommand{\cN}{\mathcal{N}}
  
  \providecommand{\cP}{\mathcal{P}}

  \providecommand{\cS}{\mathcal{S}}

\def\figref#1{Figure~\ref{#1}}
\def\figtab#1{Table~\ref{#1}}
\def\algref#1{Algorithm~\ref{#1}}

\def\bydef{\triangleq}

\newcommand{\R}{\mathbb{R}}
\DeclareMathOperator{\E}{\mathbb{E}}
\newcommand{\abs}[1]{\left\lvert #1\right\rvert}
\newcommand{\norm}[1]{\left\lVert #1\right\rVert}

\newcommand{\lin}[1]{\ensuremath \left\langle #1 \right\rangle}

\newtheorem{claim}{Claim}[section]

\theoremstyle{plain}
\newtheorem{theorem}{Theorem}[section]

\theoremstyle{definition}
\newtheorem{defn}{Definition}[section]
\newtheorem*{defn*}{Definition}

\newtheorem{corollary}{Corollary}[theorem]

\theoremstyle{remark}
\newtheorem{rem}{Remark}
\newtheorem*{rem*}{Remark}

\usepackage[suppress]{color-edits}
\addauthor{sw}{blue}
\addauthor{xc}{magenta}
\addauthor{mh}{red}

\title{{\bf Understanding Clipping for Federated Learning: Convergence and Client-Level Differential Privacy}}

 \author{Xinwei Zhang$^\dag$, Xiangyi Chen$^\dag$, Mingyi Hong$^\dag$,  Zhiwei Steven Wu$^\ddag$, and Jinfeng Yi$^\#$\\
 $^\dag$ Department of Electrical and Computer Engineering, University of Minnesota\\
 $^\ddag$ School of Computer Science, Carnegie Mellon University\\
 $^\#$ Machine Learning Department, JD.com, Inc.\\
 $^\dag$\texttt{zhan6234,chen5719,mhong@umn.edu},\\ $^\ddag$\texttt{zstevenwu@cmu.edu}, $^\#$\texttt{yijinfeng@jd.com}
 }
 
\begin{document}

\maketitle

\begin{abstract}
  Providing privacy protection has been one of the primary motivations of Federated Learning (FL). Recently, there has been a line of work on incorporating the formal privacy notion of differential privacy with FL.  To guarantee the client-level differential privacy in FL algorithms, the clients' transmitted model updates have to be {\it clipped} before {adding privacy noise}. Such clipping operation is substantially different from its counterpart of gradient clipping in the centralized differentially private SGD and has not been well-understood. In this paper, we first empirically demonstrate that the clipped FedAvg can perform surprisingly well even with substantial data heterogeneity when training neural networks, which is partly because the clients' updates become {\it similar} for several popular deep architectures. Based on this key observation,  we provide the convergence analysis of a differential private (DP) FedAvg algorithm and highlight the relationship between clipping bias and the distribution of the clients' updates.
  To the best of our knowledge, this is the first work that rigorously investigates theoretical and empirical issues regarding the clipping operation in FL algorithms.
\end{abstract}

\section{Introduction}
 First proposed by~\cite{konevcny2016federated}, \emph{Federated Learning} (FL) is a distributed learning framework that aims to reduce communication complexity and to provide privacy protection during training. The popular FedAvg algorithm \cite{konevcny2016federated} has been proposed to reduce the communication cost by using periodic averaging and client sampling. {There has been many extensions of this algorithm, mostly by modifying the local update directions \cite{karimireddy2020scaffold,Zhang_FedPD_Arxiv_2020,liang2019variance}.} Even though FL algorithms \swedit{have the goal of privacy protection}, recent works have shown that they are vulnerable to inference attacks and leak local information during training~\cite{zhao2020idlg,zhu2020deep,wei2020framework}. 
\swedit{As a result, striking a balance between \emph{formal} privacy guarantees and desirable optimization performance remains one of the fundamental challenges in FL.}

Recently, various FL algorithms~\cite{geyer2017differentially,truex2020ldp, truex2019hybrid, wang2020d2p, triastcyn2019federated} have been proposed to provide the formal guarantees of \emph{differential privacy} (DP)~\cite{Dwork2006}.
In these algorithms, the clients perform multiple local updates between two communication steps, and then perturbation mechanisms are added to aggregate updates across individual clients. In order for the perturbation mechanism to have formal privacy guarantees, each client's model update needs to have a bounded norm, which is ensured by applying a clipping operation
that shrinks individual model updates when their norm exceeds a given threshold. While there has been prior work that studies the clipping effects on stochastic gradients~\cite{bassily2014private,chen2020understanding,song2021evading} in the differentially private SGD \cite{abadi2016deep}, there has not been any work on providing understanding how clipping the model updates affect the optimization performance of FL subject to DP. Our work provides the first in-depth study on such clipping effects. 
\swdelete{Notably, the aggregation step is not directly performed on the individual stochastic gradients, which is typically the case for centralized algorithm like differentially private SGD (DP-SGD),
but instead a sequence of (stochastic) gradients. On the contrary, classical analysis for differentially private (stochastic) gradient descent directly clips the (stochastic) gradients, and the key in the analysis is {to bound the bias introduced by clipping the (stochastic) gradient} \cite{abadi2016deep}. Therefore, all such existing analyses can not be directly applied to the same operation in FL.
However, none of the existing work provides a rigorous understanding of how clipping operation affect the convergence of FL algorithms. Therefore it is not clear how these DPFL algorithms can balance between optimization performance and privacy guarantees.}

\noindent{\bf Contributions.} In this work, we will conduct rigorous theoretical analysis and provide extensive empirical evidence to understand how to best protect client-level DP for FL algorithms. Specifically, we make the following contributions: 

\noindent {\bf 1)} We analyze the existing model and difference clipping strategies for clipping-enabled FedAvg and prove that difference clipping outperforms model clipping. Our result provides theoretical insight into designing FL algorithms with clipping operation. 

\noindent {\bf 2)} We empirically show that the performance of the clipping-enabled FedAvg depends on the structure of the neural network being used -- when \swedit{the structure of the network induces \emph{concentrated} clients' updates}, and the performance drop becomes negligible.

\noindent {\bf 3)} We provide the convergence analysis of the clipping-enabled FedAvg algorithm and highlight the relationship between clipping bias and the distribution of the clients' updates. Our result leads to a natural guarantee of client-level DP for FedAvg. 

To the best of our knowledge, this is the first work that rigorously investigates theoretical and empirical issues regarding the clipping operation in FL algorithms.

\subsection{Preliminaries \& Related Work}

Federated learning typically considers the following optimization problem:
\begin{equation}\label{eq:FL}
    \min_{\xx} \bigg[f(\xx) \triangleq \sum^N_{i=1}f_i(\xx)\bigg], \;\mbox{\rm where } f_i(\xx) = \E_{\xi\sim \cD_i} F(\xx;\xi),
\end{equation}
where $N$ is the number of participating clients; the $i^\mathrm{th}$ client optimizes a local model $f_i$, which is the expectation of a loss function $F(\xx;\xi)$, where the expectation is taken over local data distribution $\cD_i$. At each communication round $t$, the server samples a subset of clients $\cP_t$ and broadcasts the global model parameters $\xx^t$. The sampled clients perform $Q$ steps of SGD updates and compute the total update differences $\Delta\xx^t_i$'s, and then the server aggregates the update differences to update the global model.  In Algorithm \ref{alg:FedAvg}, we present a slightly generalized FedAvg algorithm \cite{karimireddy2020scaffold,yang2021achieving}, in which the server uses a stepsize $\eta_g$ to perform its update. When $\eta_g=1$, the algorithm becomes the same as the original FedAvg. 

\begin{algorithm}[b!]
	\small
	\begin{algorithmic}[1]
		\STATE {Initialize: $\xx^{0}_i\bydef\xx^{0}, i=1,\dots,N$}\\
		\FOR{$t=0,\dots,T-1$ {\it (stage)}}
		\FOR{$i\in \cP_t\subseteq [N]$ in parallel}
		\STATE {Update agents' $\xx^{t,0}_i =\xx^{t}$}
		\FOR{$q=0,\dots,Q-1$ {\it (iteration)}}
		\STATE{Compute stochastic gradient  $g_i^{t,q}$ with $\mathbb E [g_i^{t,q}] = \nabla f_i(x_i^{t,q})$ 
		\STATE{Local update:} $\xx^{t,q+1}_i = \xx^{t,q}_i - \eta_l g_i^{t,q} $}
		\ENDFOR
		\ENDFOR
		\STATE {Global averaging:} { $\Delta\xx^{t}_i = \xx^{t,Q}_i - \xx^{t}$, \quad $\xx^{t+1} =\xx^{t}+\eta_g\frac{1}{|\cP_t|}\sum_{i\in\cP_t}
		\Delta\xx^{t}_i$}
		\ENDFOR
	\end{algorithmic}
	\caption{FedAvg Algorithm}\label{alg:FedAvg}
\end{algorithm}

\swedit{In this work, we study FL subject to the rigorous privacy guarantees of \emph{Differential Privacy} (DP) \cite{Dwork2006}}, whose formal definition is given below.

\begin{defn}\label{def:CDP}\cite{Dwork2006}
	An algorithm $\cM$ is $(\epsilon, \delta)$-differentially private if 
	\begin{equation}
	    P(\cM(\cD)\in\cS) \leq e^\epsilon P(\cM(\cD')\in\cS) + \delta,
	\end{equation}
	where $\cD$ and $\cD'$ \swedit{are neighboring datasets, $\cS$ is an arbitrary subset of outputs of $\cM$.} 
\end{defn}

 The common mechanism used to protect DP in centralized training is straightforward: 1) clip the stochastic gradient with the so-called clipping operation~\eqref{eq:clip}; 2) add a random perturbation $\zz\sim\cN(0,\sigma^2I)$ to the clipped quantity~\cite{abadi2016deep}. The clipping operation is the key step to guarantee DP as the noise level $\sigma^2$ is determined by the clipping threshold $c$~\cite{dwork2014algorithmic}:
\begin{equation}\label{eq:clip}
	\mbox{\rm clip}(g^{t},c) = g^t\cdot\min\bigg\{1,\frac{c}{\norm{g^t}}\bigg\}.
\end{equation}

However, DP is more complex in FL than that in centralized training. Two key factors distinguish FL from existing DP machine learning framework are:
\vspace{-0.2cm}
\begin{itemize}[leftmargin=*]
    \item {\it Data distribution}: unlike centralized training, in FL the data are naturally distributed on the clients, and the clients can potentially have very different data distributions. In the centralized setting,  the recent work  \cite{chen2020understanding} has shown that the distribution of the samples affects the performance of the DP-SGD, but how heterogeneous data distribution affects the design and analysis of FL algorithm that protects DP is unclear. 
    \item {\it Local updates}: as described in \algref{alg:FedAvg}, the clients will perform multiple local update steps before sending the model to the server, and it is well-known that when $Q>1$, the data heterogeneity will cause performance degradation in FedAvg even without clipping and perturbation \cite{khaled2019first}. Although there are multiple alternatives of how the DP mechanism can be applied to FL algorithms, none of those mechanisms has a rigorous theoretical guarantee, and it is not clear how to properly balance the optimization performance and privacy guarantees.  
\end{itemize}

These two factors result in different {\it definitions} and {\it clipping operations} in FL. 

\underline{DP definitions in FL:} Based on the distribution pattern of the client and local datasets, two DP definitions are commonly considered in FL algorithm design: 
\vspace{-0.2cm}
\begin{itemize}[leftmargin=*]
    \item Sample-level differential privacy (SL-DP):  SL-DP directly follows the centralized DP and protects each local sample so that the server could not identify one sample from the union of all local datasets, i.e., $\cD = \bigcup^N_{i=1}\cD_i$, and $\cD,\cD'$ differ by one sample $\xi$. SL-DP fits in the cross-silo FL scenario that has a relatively small number of clients, each with a large dataset. E.g., SL-DP is used in medical image classification application to protect patients' personal information~ {\cite{choudhury2019differential}}. However, in the Google Keyboard application~\cite{hard2018federated} where each client is an application user, SL-DP that only protects one sample (i.e., an input record) will not be sufficient to protect the user's personal information.
    \item Client-level differential privacy (CL-DP): CL-DP has a stricter privacy guarantee compared with SL-DP. It requires that the server cannot identify the participation of one client by observing the output of the local updates, i.e., $\cD =\{\cD_i\}^N_{i=1}$, and  $\cD,\cD'$ differ by one dataset $\cD_i$. CL-DP is suitable for the cross-device FL scenario such as the Google Keyboard application, which has a large number of distributed clients. 
\end{itemize}

\underline{Clipping operation in FL:} Based on different DP requirements and the algorithm structures, a number of FL algorithms have been proposed which protect DP to some extent. 

To protect SL-DP, \cite{truex2019hybrid} proposes to clip and inject noise to every local update. {That is, some Gaussian noise is added to the stochastic gradients $g^{t,q}_i$ given in \algref{alg:FedAvg}.} However, as intermediate updates are kept local and private, the clipping and perturbation to the local steps appear to be unnecessary, and such operations result in significant performance degradation. {Moreover, it is not clear how such kind of operation impact other aspects of the algorithm performance (such as algorithm convergence, quality of solutions, etc.)}

To protect CL-DP, \cite{wei2020federated} proposes to clip the local models to be transmitted directly. Similarly, \cite{truex2020ldp} assumes that the model parameters are upper and lower bounded by some constant and directly apply perturbations to the local models. {However, this scheme also significantly reduces the training and test accuracy empirically and has no theoretical convergence guarantee}. Recently, \cite{geyer2017differentially} proposes to clip the difference between the input model and the output models of the FedAvg algorithm. In particular, one can replace the update directions $\Delta\xx^{t}_i$'s of line 8 in Algorithm 1 by their clipped versions as expressed below: 
\begin{equation}
	\label{eq:diffclip}
	\begin{aligned}
	\mbox{\rm clip}(\Delta\xx^{t}_i,c) = \Delta\xx^{t}_i\cdot\min\bigg\{1,\frac{c}{\norm{\Delta\xx^{t}_i}}\bigg\}, \\
	\xx^{t+1} =\xx^{t}+\eta_g\frac{1}{|\cP_t|}\sum_{i\in\cP_t}
		\mbox{\rm clip}(\Delta\xx^{t}_i,c).
	\end{aligned}
\end{equation}
It is shown that such a scheme has better numerical performance than model clipping, but no convergence proof for the algorithm is given. {Reference \cite{triastcyn2019federated} also clips the update difference and proposed Bayesian DP to measure the privacy loss and only demonstrates the numerical performance of the proposed algorithm. D2P-Fed \cite{wang2020d2p} follows the same clipping strategy and further apply compression and quantization during communication to improve communication efficiency while having DP guarantee, but its convergence guarantee only applies to the non-clipping version.}

In summary, despite extensive recent research about DP-enabled FL, there are still a number of technical challenges and open research questions in this area. First, it is not clear how various kinds of clipping operations can affect the performance of FL algorithms. Second, it is not clear how to add noise to balance the convergence of FL algorithms and its CL-DP guarantee.

\section{Clipping Issues in FL}

As discussed above, clipping is a key operation in providing DP guarantee for FL algorithms. Therefore, to design algorithms that protect DP in FL, the first step  is to understand how clipping affects the convergence performance of a FL algorithm. Towards this end, we start with analyzing two common clipping strategies, and identify their theoretical properties. {Then we provide a series of empirical studies to demonstrate how system parameters such as training models, datasets and data distributions can affect the performance of clipping-enabled FedAvg algorithm.} These empirical studies will be combined with our theoretical analysis in the next section to provide a comprehensive understanding about the optimization performance and CL-DP guarantees in FL.

\subsection{Model clipping versus Difference Clipping}

The two major clipping strategies used in protecting CL-DP for FL algorithms are {\it local model} clipping and {\it local update difference} clipping, as we describe below.
\vspace{-0.2cm}
\begin{enumerate}[leftmargin=*]
    \item {\bf Model clipping}~\cite{wei2020federated}: The clients directly clip the models sent to the server. For FedAvg algorithm, this means performing $\mbox{\rm clip}(x^{t,Q}_i,c)$. This method appears to be straightforward, but clipping the model directly results in relatively large clipping threshold, so it requires to add larger perturbation. 
    \item {\bf Difference clipping}~\cite{geyer2017differentially}: The clients clip the local update difference between the initial model and the output model according to~\eqref{eq:diffclip}. This method needs to record the initial model and to perform extra computation before clipping, but the update difference typically has smaller magnitudes than the model itself, so the clipping threshold and the perturbation can be smaller than using model clipping. Note that when $Q = 1$, the difference clipping is equivalent to the standard mini-batch gradient clipping (i.e., the DP-SGD), but in the general case where $Q>1$, their behaviors are very different.
\end{enumerate}

Below we analyze how they perform on simple quadratic problems. Our results indicate that the difference clipping strategy is more preferable, because it is less likely to have strong impact on the optimization performance. 

\begin{claim}\label{eq:claim:model}
    Given any constant clipping threshold $c$, there exists a convex quadratic problem, for which FedAvg with model clipping does not converge to the global optimal solution with any fixed $Q\geq 1$ and $\eta_l > 0$.
\end{claim}
\noindent{\bf Proof.} Given a fixed clipping threshold $c$, consider the following quadratic problem $$f(x) = \sum^3_{i=1} \frac{1}{2}(x-b_i)^2,$$ where we have $N = 3$ clients. By applying model clipping to FedAvg, one round update can be expressed as:

\begin{equation*}
    \begin{aligned}
        x^+ = \frac{1}{3}\sum^3_{i=1}\mbox{\rm clip}(\lambda x+(1-\lambda)b_i,c), \text{ where } \lambda =  (1-\eta_l)^Q \in (0,1),
    \end{aligned}
\end{equation*}
{ where $\eta_l$ is the local stepsize.}

Suppose that the algorithm converges, then we will have solution $x^+ = x = x^\infty$. This implies that
\begin{align}\label{eq:updates}
    \frac{1}{3}\sum^3_{i=1}\mbox{\rm clip}(\lambda x^\infty+(1-\lambda)b_i,c) = x^\infty.
\end{align}
Let us set $b_1 = b_2 = -0.5c, b_3 = kc$, then it is easy to verify that the optimal solution of the problem is given by $x^\star = \frac{(k-1)c}{3}>0$.
However, when $k>4$, from \eqref{eq:updates} we can see that $x^\infty \leq c$ and $x^\star >c$. Therefore, the only possibility is that $x^\infty = \frac{\lambda}{3-2\lambda}c\leq c \neq x^\star$, and this holds true for any $\lambda\in(0,1)$. So the stationary solution of FedAvg with model clipping to this problem will not converge to the original optimal solution no matter how we choose $Q$ and $\eta_l$. \hfill $\blacksquare$

\begin{claim}\label{eq:claim:difference}
    For all linear regression problem with fixed clipping threshold $c$, there exist $\eta_l$ and local update step $Q\ge 1$ such that FedAvg with difference clipping converges to the global optimal solution. Furthermore, there exist a linear regression problem such that under the same $c, \eta_l$ and $Q$, FedAvg with difference clipping converges to a better solution with smaller loss than the original FedAvg.
\end{claim}

\noindent {\bf Proof.} First, we prove that using difference clipping, FedAvg can converge to global optimal by carefully selecting $Q$ and $\eta_l$. Consider the following convex quadratic problem $$f(x) = \sum^N_{i=1} \frac{1}{2}(A_ix-b_i)^2.$$ By applying FedAvg with update difference clipping, one round of update can be expressed as:

\begin{equation*}
    \begin{aligned}
        x^+ &= x - \frac{1}{N}\sum^N_{i=1} \mbox{\rm clip}(\Lambda_i \nabla f_i(x),c), \text{ where } \Lambda_i = (I - (I - \eta_l A^T_iA_i)^Q)(A^T_iA_i)^{-1}.
    \end{aligned}
\end{equation*}

In order for the problem to converge to the original problem, it is easy to verify that the following condition has to hold: $$\sum^N_{i=1} \mbox{\rm clip}(\Lambda_i \nabla f_i(x^\star),c) = 0.$$ 

The above example can be viewed as using gradient descent to optimize a problem with the following gradient 
\begin{equation}
    \nabla f'_i(x) = \left\{\begin{array}{cc}
            \Lambda_i \nabla f_i(x) &  \norm{\Lambda_i \nabla f_i(x) } \leq c,\\
            \frac{c \Lambda_i \nabla f_i(x) }{\norm{\Lambda_i \nabla f_i(x)}}& \text{otherwise}.
        \end{array}\right.
\end{equation}
Note that in general it is hard to write down the exact local problems $f'_i$ that satisfies the above condition, but when $x\in \R$ is a scalar, $f'_i(x)$ is the Huberized loss of $\Lambda_i f_i(x)$~\cite{song2021evading} 

\begin{equation}
    f'_i(x) = \begin{cases}
    \Lambda_i f_i(x)& \mbox{if } \abs{\Lambda_i A_i(A_ix -b_i)}\leq c,\\
    c\abs{\frac{\Lambda_i}{A_i} f_i(x)} -\frac{1}{2}c^2& \mbox{otherwise.}
    \end{cases}
\end{equation}

In general, the re-weighted problem does not have the same solution as the original problem, but we can select $\eta_l$ and $Q$ (determined by on $x^\star$ and $f_i$'s) so that $f'(x)$ has the same solution as $f(x)$. For example, one set of parameters that satisfy the above requirement is $Q = 1, \eta_l = 1/\max_i\{\norm{\nabla f_i(x^\star)}\}$. In this case, $\Lambda_i = I\eta_l$, and when $\eta_l$ is small enough, the clipping will not be activate when $x = x^\star$ and $\sum^N_{i=1} \mbox{\rm clip}(\Lambda_i \nabla f_i(x^\star),c) = \sum^N_{i=1} \eta_l\nabla f_i(x^\star) = 0$. 

Next, we show that Clipping-enabled FedAvg can outperform the non-clipped version. Note that when $Q > 1$, even when $\eta$ is small such that the clipping is not activated, the algorithm will not converge to the original solution. So in general one cannot draw the conclusion about whether clipping helps or hurts the performance of FedAvg. Consider the following problem: 

\begin{equation}\label{eq:example_3}
    \begin{aligned}
        f(x) &= \sum^3_{i=1} f_i(x),\\
        f_1(x) &= \frac{1}{2}(x-4)^2,\; f_2(x) = \frac{1}{2}(2x-1)^2, \; f_3(x) = \frac{1}{2}(6x+1)^2.
    \end{aligned}
\end{equation}

As $\nabla f(x) = (x-4) + (4x-2) + (36x +6) = 41x$, the optimal solution of this problem is $x^\star = 0$. Table \ref{tab:solutions} show the stationary points of FedAvg under different choice of parameters. When $Q = 1$, FedAvg is equivalent to SGD and clipping hurts the performance of FedAvg. However, when $Q$ is large, clipped FedAvg has a better performance than the non-clipped version, in the sense that the stationary solution it obtains are closer to the global optimal solution $x^*=0$. \hfill $\blacksquare$

\begin{table}[tb!]
    \centering
    \begin{tabular}{c|cc}
    \hline
                        & $Q = 1$                   & $Q = \infty$ \\
                        \hline
        $c = \infty$    & $x^\infty = 0$            & $x^\infty = \frac{13}{9}$ \\
        $c = 1$         & $x^\infty = \frac{1}{2}$  & $x^\infty = \frac{2}{3}$\\
        \hline
    \end{tabular}
    \caption{Stationary points of FedAvg with gradient clipping for \eqref{eq:example_3} under different parameter settings.}
    \vspace{-0.8cm}
    \label{tab:solutions}
\end{table}

\begin{rem}
To prove Claim \ref{eq:claim:model}, we construct a problem whose magnitude of the optimal solution is larger than the clipping threshold. Then FedAvg with model clipping will converge to a stationary point with magnitude bounded by the clipping threshold, therefore the algorithm will not converge to global optimal solution.

The technique to prove the first part of Claim \ref{eq:claim:difference} is related to the analysis for centralized gradient clipping algorithms \cite{song2020characterizing}. The main difference is that our algorithm considers $Q$ steps of local update before clipping. We show that by allowing multiple local updates, FedAvg algorithm with difference clipping optimizes the sum of the Huberzied re-weighted local loss functions. By properly choosing the learning rate $\eta_l$ for each local loss function, we can balance the re-weighting factors so that the optimal solution to the new loss function matches the solution to the original problem.  \hfill $\blacksquare$
\end{rem}

The above claims indicate that the difference clipping should outperform the model clipping in terms of convergence guarantees. Therefore, in the subsequent analysis, we will focus on understanding the difference clipping enabled FL algorithms. In particular, we consider the Clipping-Enabled FedAvg (CE-FedAvg) algorithm described in~\algref{alg:CE-FedAvg}, which combines the difference clipping with the slightly generalized FedAvg algorithm described in \algref{alg:FedAvg} (which uses two stepsizes $\eta_l, \eta_g$, one for local and one for global updates, respectively). The reason to consider such a {\it bi-level-stepsize} version of FedAvg is that,  it has been proved to have superior performance, especially when not all clients participate in each round of communication~\cite{karimireddy2020scaffold,yang2021achieving}. 


\begin{algorithm}[b!]
	\small
	\begin{algorithmic}[1]
		\STATE {Initialize: $\xx^{0}_i\bydef\xx^{0}, i=1,\dots,N$}\\
		\FOR{$t=0,\dots,T-1$ {\it (stage)}}
		\FOR{$i\in \cP_t\subseteq [N]$ in parallel}
		\STATE {Update agents' $\xx^{t,0}_i =\xx^{t}$}
		\FOR{$q=0,\dots,Q-1$ {\it (iteration)}}
		\STATE{Compute stochastic gradient  $g_i^{t,q}$ with $\mathbb E [g_i^{t,q}] = \nabla f_i(x_i^{t,q})$ 
		\STATE{Local update:} $\xx^{t,q+1}_i = \xx^{t,q}_i - \eta_l g_i^{t,q} $}
		\ENDFOR
		\STATE{Compute update difference:} $\Delta\xx^{t}_i = \xx^{t,Q}_i - \xx^{t,0}_i$
		\STATE{Clip:} $\Hat{\Delta}\xx^{t}_i = {\rm clip}(\Delta\xx^t_i,c)$, where ${\rm clip}(\cdot)$ is defined in \eqref{eq:clip}
		\ENDFOR
		\STATE {Global averaging:} { $\xx^{t+1} =\xx^{t}+\eta_g\frac{1}{|\cP_t|}\sum_{i\in\cP_t}\Hat{\Delta}\xx^{t}_i$}
		\ENDFOR
	\end{algorithmic}
	\caption{Clipping-enabled FedAvg Algorithm (CE-FedAvg)}\label{alg:CE-FedAvg}
\end{algorithm}

\subsection{Empirical Results}

\noindent{\bf Experiment Setting.} To have a thorough understanding about how the difference clipping can impact the FedAvg, we conduct numerical experiments with different models, datasets and local data distributions. We compare the test accuracies between CE-FedAvg and the original FedAvg.  Note that in this set of experiments we do not consider the privacy issues yet, so we do not add perturbation.

To have a fair comparison, we set $Q$, $T$, $N$, $\abs{\cP_t}$, $\eta_l$ and $\eta_g$ to be identical for both FedAvg and CE-FedAvg. We first run the original FedAvg, compute $\norm{\Delta\xx^{t}_i}$ and average over all clients $i$ and iterations $t$ to obtain $\bar{\Delta}$ and choose the clipping threshold $c=0.5\bar{\Delta}$. 

We run the algorithm using AlexNet~\cite{krizhevsky2012imagenet} and ResNet-18~\cite{he2016deep} with EMNIST dataset~\cite{cohen2017emnist} and Cifar-10 dataset~\cite{krizhevsky2009learning} for comparison. We split the dataset in two different ways: 1) {\it IID Data} setting, where the samples are uniformly distributed to each client; 2) {\it Non-IID Data} setting, where the clients have unbalanced samples. Details are described below. For EMNIST digit classification dataset, each client has 500 samples without overlapping. In the IID case, each client has around 50 samples of each class and in the Non-IID case, there are 8 classes each has around 5 samples and 2 classes each has 230 samples on each client. For the Cifar-10 dataset, in the IID case (resp. Non-IID case), each client also has 500 samples (resp. $50$ samples); these samples can overlap with those on the other clients and {the samples on each client are uniformly distributed in $10$ classes, i.e., each client has $50$ samples (resp. $5$ samples) from each class.}

\noindent{\bf Performance Degradation.} In \figtab{tab:clip_drop}, we compare the classification results produced by using AlexNet and ResNet-18 on the two datasets.
{\begin{table*}[t!]
    \centering\footnotesize
    \begin{tabular}{l|rrrrr}
        \hline
         Model& dataset& IID(\%)& IID Clipping (\% drop) & Non-IID (\%)& Non-IID Clipping (\% drop)\\
         \hline
         AlexNet 	& EMNIST	& 98.20	& 0.19	& 95.60 & 3.60\\
		  			& Cifar-10	& 66.01	& 4.83	& 57.14 & 7.30\\
         ResNet-18 	& EMNIST	& 99.61 & 0.02 	& 95.43 & 0.10\\
		 			& Cifar-10	& 76.36	& 0.53	& 59.46 & 1.55\\
         \hline
    \end{tabular}
    \caption{\footnotesize The accuracy drop between a) FedAvg and clipping-enabled FedAvg, used for training AlexNet and ResNet-18, on IID and Non-IID data.}
    \vspace{-0.8cm}
    \label{tab:clip_drop}
\end{table*}}

There are three interesting observations: 1) The data distribution will greatly affect the clipping performance in FL. When data are IID across the clients, clipping has far less impact on the final accuracy, otherwise the clipping will introduce some accuracy drop to the trained models; 2) Clipping has quite different impact on different models -- the best accuracy of the models drops $0.10\%$  and  $3.60\%$ for ResNet-18 and AlexNet on EMNIST, respetively. The drop is $1.55\%$ for ResNet-18 and $7.30\%$ {for AlexNet on Cifar-10}, comparing CE-FedAvg with non-clipped version on the Non-IID data; 3) Data complexity also affects the behavior of the CE-FedAvg -- the accuracy drop on Cifar-10 dataset is much larger than that on EMNIST dataset.

The empirical experiments show that heterogeneous data distribution among the clients is one of the main causes of the different behavior between the clipped and non-clipped algorithms. The data heterogeneity issue is unique in FL cause by periodical communication. It does not happen in centralized optimization where the data are shared among all workers. 

\begin{figure}[t!]
    \centering
     \begin{subfigure}[b]{0.24\textwidth}
         \centering         
		 \includegraphics[width=\textwidth]{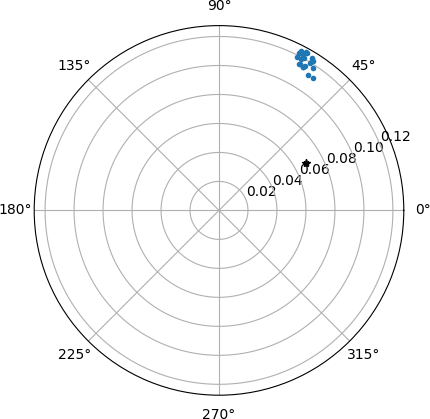}
         \caption{AlexNet, IID}
         \label{fig:alexiid}
     \end{subfigure}
     \hfill
     \begin{subfigure}[b]{0.24\textwidth}
         \centering
         \includegraphics[width=\textwidth]{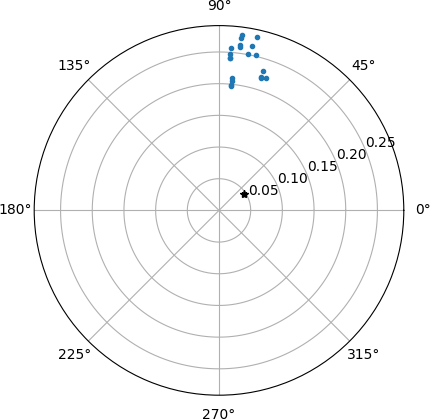}
         \caption{AlexNet, Non-IID}
         \label{fig:alexnoniid}
     \end{subfigure}
     \begin{subfigure}[b]{0.24\textwidth}
         \centering
         \includegraphics[width=\textwidth]{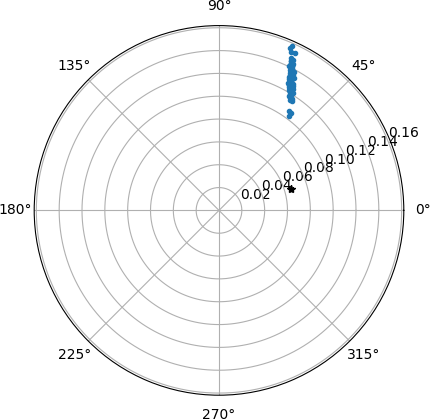}
         \caption{ResNet-18, IID}
         \label{fig:resnetiid}
     \end{subfigure}
     \hfill
     \begin{subfigure}[b]{0.24\textwidth}
         \centering
         \includegraphics[width=\textwidth]{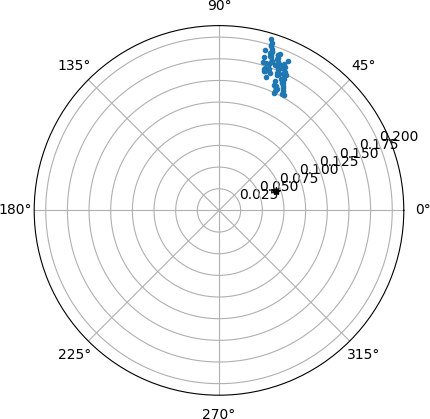}
         \caption{ResNet-18, Non-IID}
         \label{fig:resnetnoniid}
     \end{subfigure}
    \caption{\footnotesize The distribution of local updates for AlexNet and ResNet-18 on IID and Non-IID data at communication round $16$ for EMNIST dataset. Each blue dot corresponds to the local update from one client. The black dot shows the magnitude and the cosine angle of averaged local update at iteration $t$.}
    \label{fig:distribution1}
\end{figure}

\begin{figure}[t!]
		\centering
		 \begin{subfigure}[b]{0.24\textwidth}
			 \centering         
			 \includegraphics[width=\textwidth]{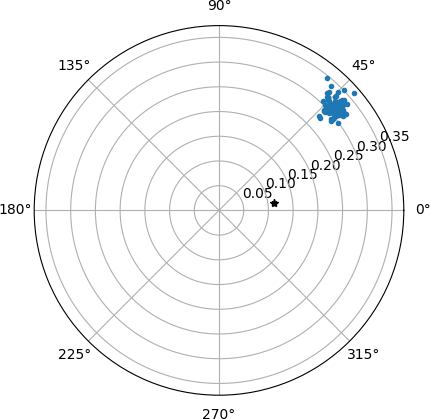}
			 \caption{AlexNet, IID}
			 \label{fig:alexiidc}
		 \end{subfigure}
		 \hfill
		 \begin{subfigure}[b]{0.24\textwidth}
			 \centering
			 \includegraphics[width=\textwidth]{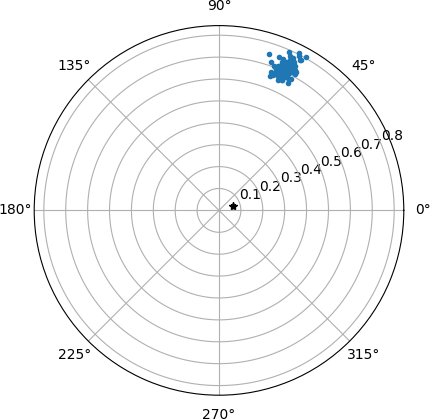}
			 \caption{AlexNet, Non-IID}
			 \label{fig:alexnoniidc}
		 \end{subfigure}
		 \begin{subfigure}[b]{0.24\textwidth}
			 \centering
			 \includegraphics[width=\textwidth]{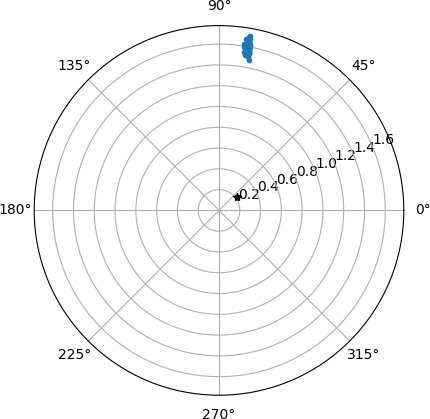}
			 \caption{ResNet-18, IID}
			 \label{fig:resnetiidc}
		 \end{subfigure}
		 \hfill
		 \begin{subfigure}[b]{0.24\textwidth}
			 \centering
			 \includegraphics[width=\textwidth]{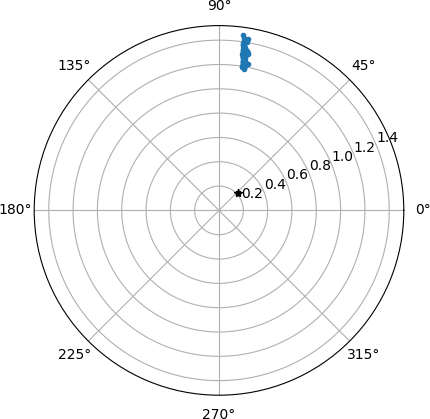}
			 \caption{ResNet-18, Non-IID}
			 \label{fig:resnetnoniidc}
		 \end{subfigure}
		\caption{\footnotesize The distribution of local updates for AlexNet and ResNet-18 on IID and Non-IID data at communication round $16$ for Cifar-10 dataset. Each blue dot corresponds to the local update from one client. The black dot shows the magnitude and the cosine angle of averaged local update at iteration $t$.}
		\label{fig:distribution2}
		\vspace{-0.7cm}
	\end{figure}

\noindent{\bf Update Difference Distribution.} To further understand the clipping procedure, we plot in Fig. \ref{fig:distribution1} and Fig. \ref{fig:distribution2} the magnitudes of local updates $\norm{\Delta\xx^t_i}$ and the cosine angles between the last iteration's global update and $\Delta\xx^t_i$:
\xcdelete{of each client:} $\cos^{-1}\left({\lin{\Delta\xx^t_i,\frac{1}{|\cP_t|}\sum_{i\in\cP_{t-1}}{\Delta}\xx^{t-1}_i}}/{\norm{\Delta\xx^t_i}\norm{\frac{1}{|\cP_t|}\sum_{i\in\cP_{t-1}}{\Delta}\xx^{t-1}_i}}\right).$ Due to page limitation, we only put the distribution of communication round $T = 16$. More detailed results are given in Appendix \ref{app:exp}. {In the plots, we mainly focus on the variance of the magnitudes of the clients' update difference (i.e., the blue dots). Larger variance indicates that the updates made by different clients are more different from each other.}

Compare Fig. \ref{fig:distribution1} with Fig. \ref{fig:distribution2} we can see that the update magnitudes on EMNIST dataset are more concentrated than that on Cifar-10 dataset by having smaller mean and variance. Similarly, by comparing Fig. \ref{fig:alexiid} with Fig. \ref{fig:alexnoniid} or Fig. \ref{fig:resnetiid} with Fig. \ref{fig:resnetnoniid}, it is clear that the local update magnitudes are more concentrated on IID data than on Non-IID data. Moreover, ResNet-18 has a more concentrated distribution of update magnitudes than AlexNet. Importantly, comparing Table \ref{tab:clip_drop} with Fig. \ref{fig:distribution1} and Fig. \ref{fig:distribution2}, one can observe that the drop in final accuracy of a model caused by clipping is correlated with {\it the degree of concentration} of update magnitudes, as AlexNet with less concentrated update magnitudes suffers more from clipping, while ResNet-18 exhibits the opposite behavior.

The above results about the update difference distributions match the accuracy results in \figtab{tab:clip_drop}, in the sense that clipping performs worse when update differences distribution has a larger divergence and vise versa. Inspired by this observation, in the next subsection, we will characterize the impact of clipping based on the degree of concentration in local updates and develop the convergence analysis of CE-FedAvg.

\section{Convergence Analysis of Clipping-Enabled FedAvg}

In this section, we analyze the theoretical performance of CE-FedAvg as well as its randomly perturbed version, in order to gain a better understanding of our previous empirical observations and the trad-off between the convergence performance of FedAvg and its DP guarantees.

Towards this end, we will provide the convergence analysis and privacy guarantees for the DP-FedAvg algorithm, described in in \algref{alg:DP-FedAvg}. Compared to CE-FedAvg, this algorithm further adds a random perturbation $\zz^t_i$ to the locally clipped model differences. 
During the communication, we assume that the attacker can only observe the aggregated update $\sum_{i\in\cP_t}\Tilde\Delta\xx^{t}_i$, and this can be guaranteed by using secure aggregation~\cite{bonawitz2017practical} or assuming the uplink of the clients to the server is secure.

Despite the similar mechanism used in DPSGD and DP-FedAvg, let us point their major differences: in DPSGD, the goal is to protect SL-DP, while DP-FedAvg is to protect CL-DP. The key difference in DP-FedAvg is that the local dataset size is large enough so that after performing multiple local update steps, the resulting model has relatively good performance. By doing so, we can largely reduce the number of communication and the corresponding privacy noise added per communication. Note that DP-FedAvg becomes DPSGD with the following choices of hyperparameters: 1) enlarge the client number to be the same as the size of the dataset, 2) decrease the local dataset size to 1; 3) decrease the number of local update to 1; 4) decrease the privacy noise accordingly.

\begin{algorithm}[b!]
	\small
	\begin{algorithmic}[1]
		\STATE {Initialize: $\xx^{0}_i\bydef\xx^{0}, i=1,\dots,N$}\\
		\FOR{$t=0,\dots,T-1$ {\it (stage)}}
		\FOR{$i\in \cP_t\subseteq [N]$ in parallel}
		\STATE {Update agents' $\xx^{t,0}_i =\xx^{t}$}
		\FOR{$q=0,\dots,Q-1$ {\it (iteration)}}
		\STATE{Compute stochastic gradient  $g_i^{t,q}$ with $\mathbb E [g_i^{t,q}] = \nabla f_i(x_i^{t,q})$ 
		\STATE{Local update:} $\xx^{t,q+1}_i = \xx^{t,q}_i - \eta_l g_i^{t,q} $}
		\ENDFOR
		\STATE{Compute update difference:} $\Delta\xx^{t}_i = \xx^{t,Q}_i - \xx^{t,0}_i$
		\STATE{Clip and perturb:} $\Tilde{\Delta}\xx^{t}_i = {\rm clip}(\Delta\xx^t_i,c)+\zz^t_i$, where ${\rm clip}(\cdot)$ is defined in \eqref{eq:clip}
		\ENDFOR
		\STATE {Global averaging:} { $\xx^{t+1} =\xx^{t}+\eta_g\frac{1}{|\cP_t|}\sum_{i\in\cP_t}\Tilde\Delta\xx^{t}_i$}
		\ENDFOR
	\end{algorithmic}
	\caption{DP-FedAvg Algorithm}\label{alg:DP-FedAvg}
\end{algorithm}

\subsection{Convergence Analysis}

\begin{theorem}[Convergence of DP-FedAvg]\label{thm: fedavg_conv}
For Algorithm \ref{alg:DP-FedAvg}, assume
\begin{align*}
   \|\nabla f_i(x) - \nabla f_i(y)\| & \leq L \|x-y\|,\; \forall\; i,x,y,\quad \min_x f(x) \geq f^*;\\
   \mathbb E [\|g_i^{t,q} - \nabla f_i(x_i^{t,q})\|^2] & \leq \sigma_l^2, \;\quad\|g_i^{t,q}\| \leq G, \; \forall\; t,q,i,\quad \|\nabla f_i(x) - \nabla f(x)\|^2 & \leq \sigma_g^2, \; \forall i,
\end{align*}
where $L$ is the Lipschitz constant of gradient, $\sigma_l^2$ and $\sigma_g^2$ are intra-client and inter-client  gradient variance, $G$ is the bound on stochastic gradient.     

By letting $\eta_g\eta_l \leq \min\{\frac{P}{48Q},\frac{P}{6QL(P-1)}\}$ and $\eta_l \leq \frac{1}{\sqrt{60}QL}$, we have
{\small
\begin{align*}
	& \frac{1}{T} \sum_{t=1}^T \mathbb E [\overline \alpha^t \|\nabla f(x^t)\|^2]  \\
	&\leq  \underbrace{\frac{4(f(x^0) - f^*)}{\eta_g \eta_l QT} +  \frac{25}{2}\eta_l^2LQ(\sigma_l^2 + 6Q\sigma_g^2) \gamma_1(T) +\frac{6\eta_g \eta_l  L \sigma_l^2}{P} \gamma_2(T) }_{\text{\rm standard terms for FedAvg}} +\underbrace{ \frac{2\eta_gLd \sigma^2}{\eta_lPQ}}_{\text{\rm caused by privacy noise}} \nonumber \\ 
	& + \underbrace{G^2\frac{4}{T} \sum_{t=1}^T \mathbb E\left[\frac{1}{N} \sum_{i=1}^N (|\alpha_i^t - \tilde \alpha_i^t| + |\tilde \alpha_i^t - \overline \alpha^t|)\right]}_{\text{\rm caused by clipping}}   + \underbrace{\eta_g\eta_lLQG^2 \frac{6}{T} \sum_{t=1}^T \mathbb E\left[ \frac{1}{P} \sum_{i=1}^N (|\alpha_i^t - \tilde \alpha_i^t|^2 + |\tilde \alpha_i^t - \overline \alpha^t|^2)\right]}_{\text{\rm caused by clipping}}
\end{align*}
}%
where $P := \abs{\cP_t}$, $\alpha_i^t := \frac{c}{\max(c, \eta_l\|\sum_{q=0}^{Q-1} g_i^{t,q}\|)}$, $\tilde \alpha_i^t := \frac{c}{\max(c, \eta_l \| \mathbb E [\sum_{q=0}^{Q-1} g_i^{t,q}]\|)}$, $\overline \alpha^t := \frac{1}{N} \sum_{i=1}^N \tilde \alpha_i^t$; $d$ is the dimension of $x$, $\gamma_1(T) =\frac{1}{T}\sum_{t=1}^T \mathbb E [\overline \alpha^t] \leq 1 $, $\gamma_2(T) =\frac{1}{T}\sum_{t=1}^T \mathbb E [(\overline \alpha^t)^2] \leq 1$. 
\end{theorem}

In the bound of Theorem \ref{thm: fedavg_conv}, the standard terms are inherited from standard FedAvg with two-sided learning rates which can yield a convergence rate of $O(\frac{1}{\sqrt{PQT}}+\frac{1}{T})$ when setting $\eta_g=\sqrt{QP}$ and $\eta_l=\frac{1}{\sqrt{T}QL}$. When there is no clipping bias and privacy noise, Theorem \ref{thm: fedavg_conv} exactly recovers the standard convergence bounds for FedAvg up to a constant, see  
Theorem 1  in \cite{yang2021achieving}. In addition to the standard terms, we have extra terms caused by the privacy noise $z_i^{t}$ and the clipping operation. We highlight the terms caused by clipping which characterize the \swdelete{worse-case}\mhdelete{adversarial effect of} estimation bias caused by clipping. The bias can be decomposed into terms caused by $|\alpha_i^t - \tilde \alpha_i^t|$ and terms caused by $|\tilde \alpha_i^t - \overline \alpha^t|$. \xcedit{Notice that  $|\alpha_i^t - \tilde \alpha_i^t| \leq \eta_l |\|\sum_{q=0}^{Q-1} g_i^{t,q}\| -  \| \mathbb E [\sum_{q=0}^{Q-1} g_i^{t,q}]\||$ } , it  is clear \mhdelete{we will have} $\mathbb E [|\alpha_i^t - \tilde \alpha_i^t|] $ will be small if the stochastic local updates have similar \xcedit{variance or} magnitudes in norm, and $\mathbb E [|\alpha_i^t - \tilde \alpha_i^t|] =0$ if $\sigma_l=0$. \xcedit{This term characterizes the bias caused by local update variance.} In addition,  $\mathbb E[|\tilde \alpha_i^t - \overline \alpha^t|]$ will be small if the expected local model updates have similar magnitudes in norm across clients and $\mathbb E[|\tilde \alpha_i^t - \overline \alpha^t|]= 0$ if $ \|\mathbb E [ \Delta x^t_i]\| =  \|\mathbb E [\Delta x^t_j]\|, \forall i, j$. \xcedit{This term shows the bias caused by cross-client update variance.}

In FL, sometimes each client will have limited amount of data, and the local model updates can be performed with small $\sigma_l$ or even $\sigma_l=0$ (full batch update). Thus, the bias caused by $|\alpha_i^t - \tilde \alpha_i^t|$ can be small \mhedit{and is} avoidable. However, the bias caused by $|\tilde \alpha_i^t - \overline \alpha^t|$ is {\it unavoidable} since this term will not diminish even each client updates its local model with full batch gradient. In addition, this term might be large with heterogeneous data distribution since the heterogeneity may induce quite disparate gradient distributions across clients. Thus, it is crucial to investigate the bias caused by $|\tilde \alpha_i^t - \overline \alpha^t|$ in practice. Note that $|\tilde \alpha_i^t - \overline \alpha^t|$ is fully controlled by  differences in magnitudes of local model updates when $\sigma_l = 0$ for fixed $c$. Going back to Fig. \ref{fig:distribution1}, we do see that how such differences in update magnitudes can be affected by both the neural network models and data heterogeneity. 


\subsection{Differential Privacy Guarantee}

The privacy guarantee of DP-FedAvg can be characterized by standard privacy theorems on Gaussian mechanism. We rephrase \cite[Theorem 1 ]{abadi2016deep} for client privacy in Theorem \ref{thm: privacy}.  
\begin{theorem}[Privacy of DP-FedAvg]\label{thm: privacy}
There exist constants $u$ and $v$ so that
given the number of iterations $T$, for any $\epsilon \leq u q^2T$ with  $q= \frac{P}{N}$ and $|\mathcal P_t|= P,\  \forall t$, Algorithm \ref{alg:FedAvg} is  $(\epsilon,\delta)$-differentially private for any $\delta >0$ if $\sigma^2 \geq v\frac{c^2 PT\ln(\frac{1}{\delta})}{N^2\epsilon^2}$.
\end{theorem}
The privacy-utility trade-off of DP-FedAvg can be analyzed by substituting $\sigma^2$ from Theorem \ref{thm: privacy} into Theorem \ref{thm: fedavg_conv}.  To get more insights on how parameters like $T, \eta_g, \eta_l$ and $\epsilon$ affect DP-FedAvg, let us consider simplified Theorem \ref{thm: fedavg_conv} in Corollary \ref{corl: nobias} with $c \geq \eta_lQG$ and $\sigma^2$ substituted . If $c' < G$ in Corollary \ref{corl: nobias}, then there will be extra bias terms inherited from the bound in Theorem \ref{thm: fedavg_conv}. \mhdelete{it can be affected by $c'$ and the distribution of update magnitude of different clients.}
\begin{corollary}[Convergence with privacy guarantee]\label{corl: nobias}
Assume all assumptions in Theorem \ref{thm: fedavg_conv}, for any clipping threshold $c = \eta_l Q c' $ with $c' \geq G$, and set $\sigma^2$ as in Theorem \ref{thm: privacy}, for any $(\epsilon, \delta)$ satisfying the constraints in Theorem \ref{thm: privacy}, we have
{\small
\begin{align}\label{eq: convergence}
	& \frac{1}{T} \sum_{t=1}^T \mathbb E [ \|\nabla f(x^t)\|^2]  \leq  \underbrace{O\left(\frac{1}{\eta_g \eta_l QT} +  \eta_l^2Q^2  +\frac{\eta_g \eta_l }{P}\right)  }_{\text{\rm standard terms for FedAvg}} +\underbrace{O\left( \frac{\eta_g\eta_lQTd  \ln(\frac{1}{\delta})}{N^2\epsilon^2}\right)}_{\text{\rm caused by privacy noise}} 
\end{align}}
and the best rate one can get from the above bound is $\tilde O(\frac{\sqrt{d}}{N\epsilon})$  by  optimizing $\eta_g, \eta_l,Q,T$. 
\end{corollary}
A direct implication of Corollary \ref{corl: nobias} is that the big-$O$ convergence rate of DP-FedAvg is the same as differentially private SGD (DP-SGD) in terms of $d$, $\epsilon$, and $N$ (note that $N$ which will be number of training samples in DP-SGD).

\section{Numerical Experiments}

In the experiment, we compare the performance of FedAvg, CE-FedAvg and DP-FedAvg on two datasets. In both experiments, we set client number $N = 1920$, the number of client participates in each round {$\abs{\cP_t} = 80,\ \forall~t$}, the number of local iterations $Q = 32$ and the mini-batch size $64$. The clipping threshold is set to $50\%$ of the average \xcedit{(over clients and iterations)} of local update magnitudes recorded in FedAvg. For DP-FedAvg we set the clipping threshold the same as in CE-FedAvg, we fix the number of communication rounds and privacy budget for the algorithms to obtain the noise variance that needs to be added. Among all the experiments, we fix privacy budget $\delta = 10^{-5}.$ 

\begin{figure}[t!]
 \vspace{-0.5cm}
    \centering
     \begin{subfigure}[b]{0.45\textwidth}
         \centering
         \includegraphics[width=\textwidth]{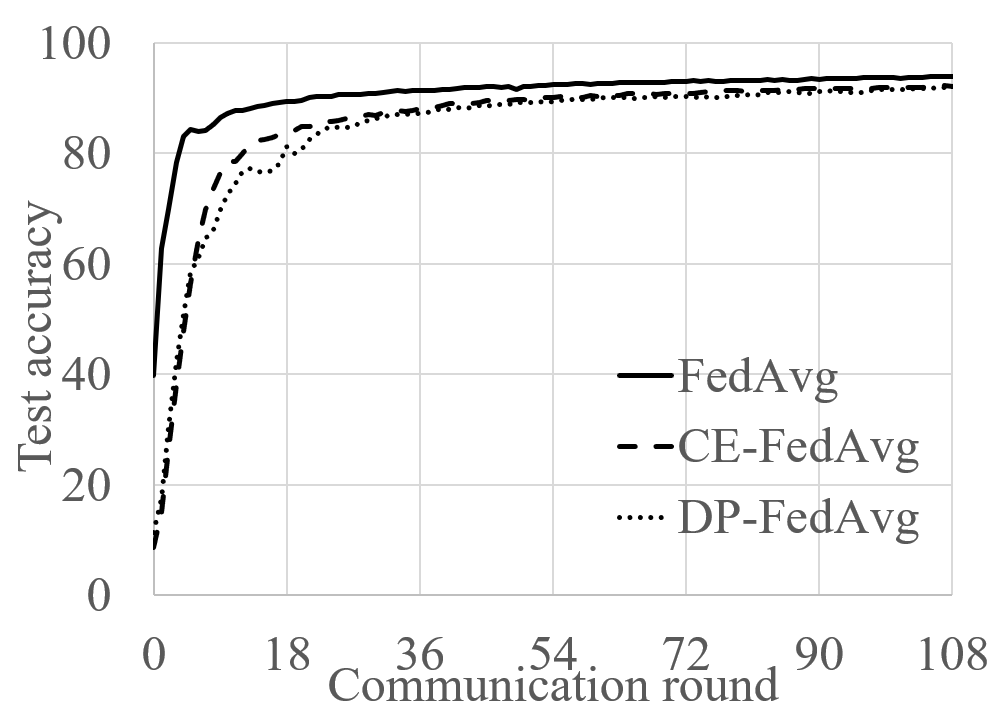}
         \caption{MLP, $\epsilon = 1.5$}
     \end{subfigure}
     \hfill
     \begin{subfigure}[b]{0.45\textwidth}
         \centering
         \includegraphics[width=\textwidth]{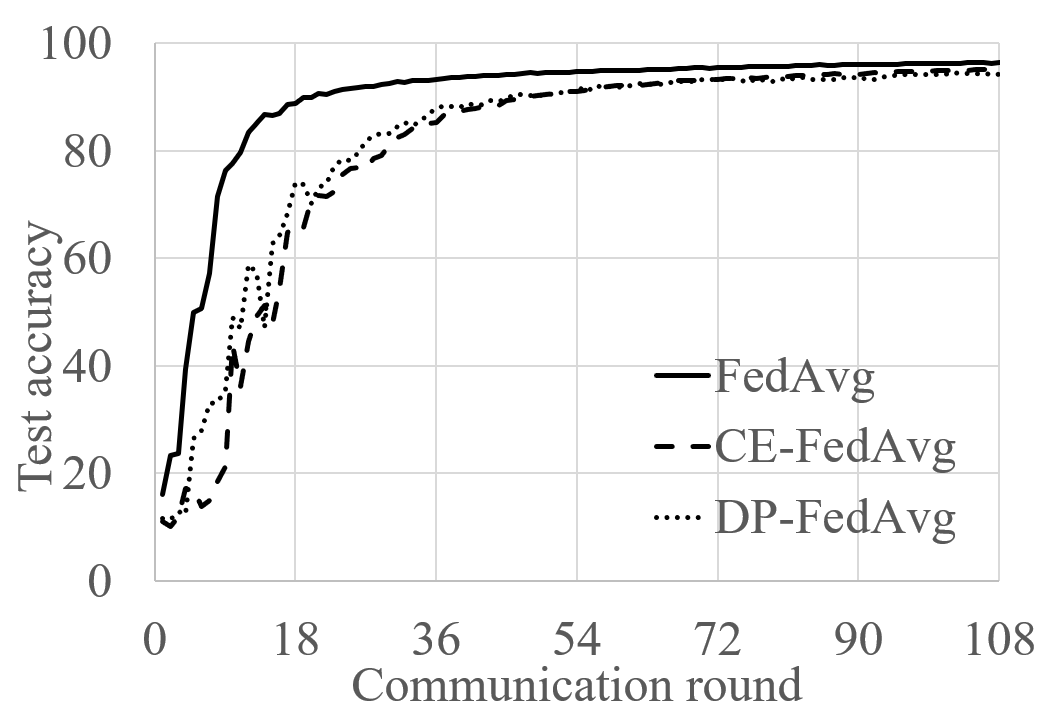}
         \caption{AlexNet, $\epsilon = 1.5$}
     \end{subfigure}
     \begin{subfigure}[b]{0.45\textwidth}
         \centering
         \includegraphics[width=\textwidth]{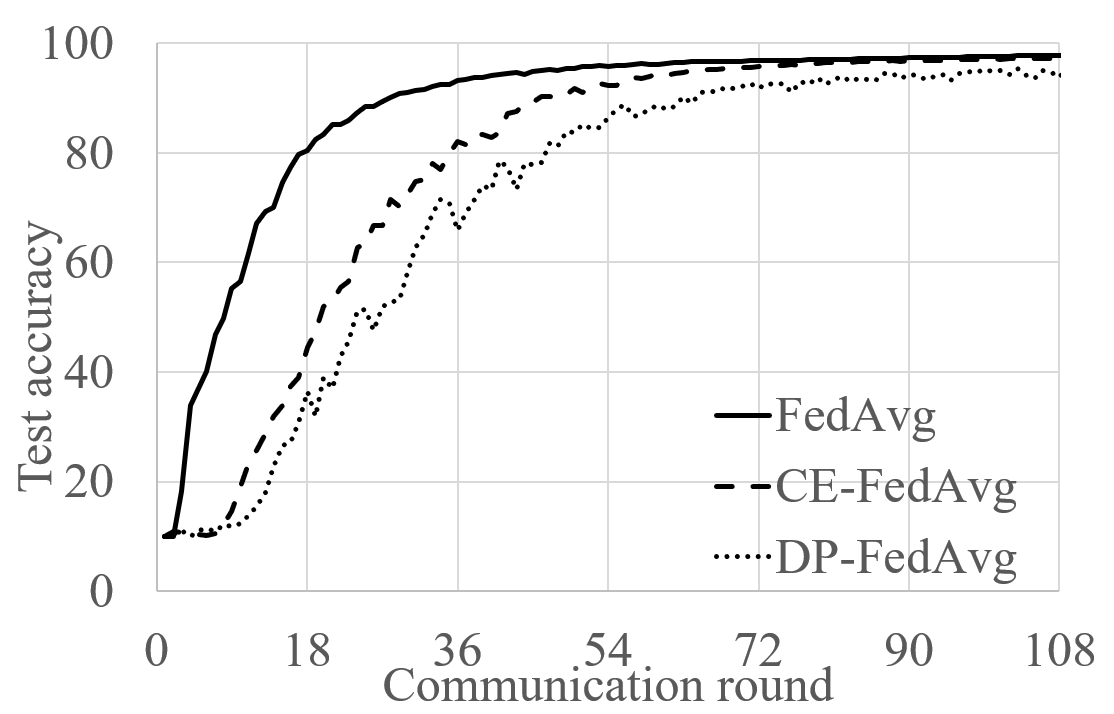}
         \caption{MobileNetV2, $\epsilon = 1.5$}
     \end{subfigure}
     \hfill
     \begin{subfigure}[b]{0.45\textwidth}
         \centering
         \includegraphics[width=\textwidth]{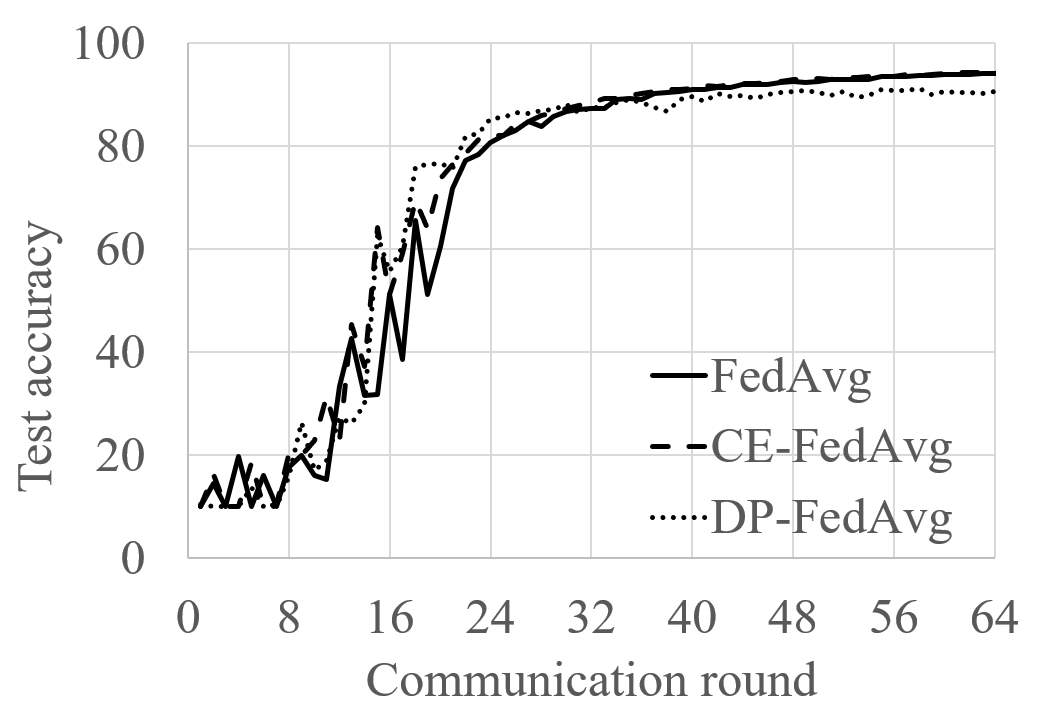}
         \caption{ResNet-18, $\epsilon = 5$}
     \end{subfigure}
    \caption{\footnotesize The test accuracy of FedAvg, CE-FedAvg and DP-FedAvg on different models on EMNIST. The privacy budgets for MLP, AlexNet and MobileNet are $\epsilon = 1.5$ while for ResNet, we set $\epsilon = 5$.}
    \label{fig:test_acc_dp1}
    \vspace{-0.5cm}
\end{figure}

\noindent{\bf EMNIST dataset.} We use the digit part of the EMNIST dataset, which has 240K training samples and 40K testing samples. We distribute the data in the Non-IID way described in Section II and each client has 125 samples. We conduct experiments on a 2-layer MLP with one hidden layer, AlexNet, ModelNetV2~\cite{sandler2018mobilenetv2} and ResNet-18. The results are listed in Table~\ref{tab:performance1} and \figref{fig:test_acc_dp1}.

{\begin{table}[b!]
 \vspace{-0.5cm}
	\centering
	\begin{tabular}{l|rrrrr}
		\hline
			Model&  \# Parameters & \# Layers & Accuracy (\%)& Clipping (\% drop) & DP (\% drop)\\
			\hline
			MLP& 159K & 2 & 94.0 & 1.84 & 0.29\\
			AlexNet & 3.3M & 7 & 96.4 & 1.47 & 0.16\\
			MobileNetV2 & 2.3M & 24 & 97.8 & 0.35 & 1.62\\
			ResNet-18 & 11.1M & 18 & 95.2 & -0.15 & 3.76$^*$ \\
			\hline
	\end{tabular}
	\caption{\footnotesize The accuracy drop between a) FedAvg and clip-enabled FedAvg and b) clip-enabled FedAvg and DP-FedAvg. The clipping threshold is $0.5$ of the average magnitude and privacy budget $\epsilon = 1.5$ for MLP, AlexNet and MobileNetV2 and $\epsilon = 5$ for ResNet-18.}
	\label{tab:performance1}
\vspace{-0.5cm}
\end{table}}

\begin{figure}[t!]
    \centering
     \begin{subfigure}[b]{0.30\textwidth}
         \centering
         \includegraphics[width=\textwidth]{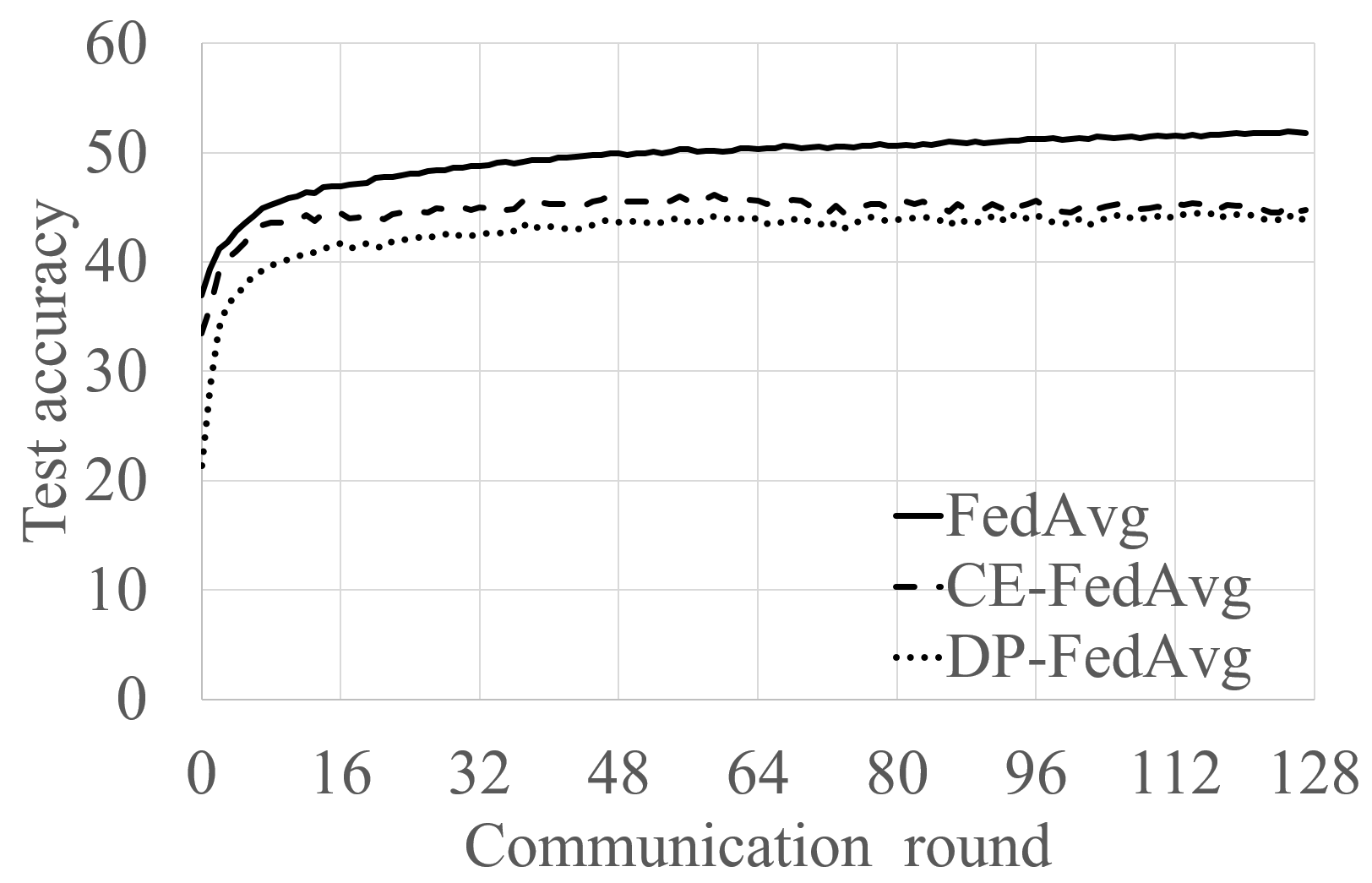}
         \caption{MLP, $\epsilon = 1.5$}
     \end{subfigure}
     \hfill
     \begin{subfigure}[b]{0.30\textwidth}
         \centering
         \includegraphics[width=\textwidth]{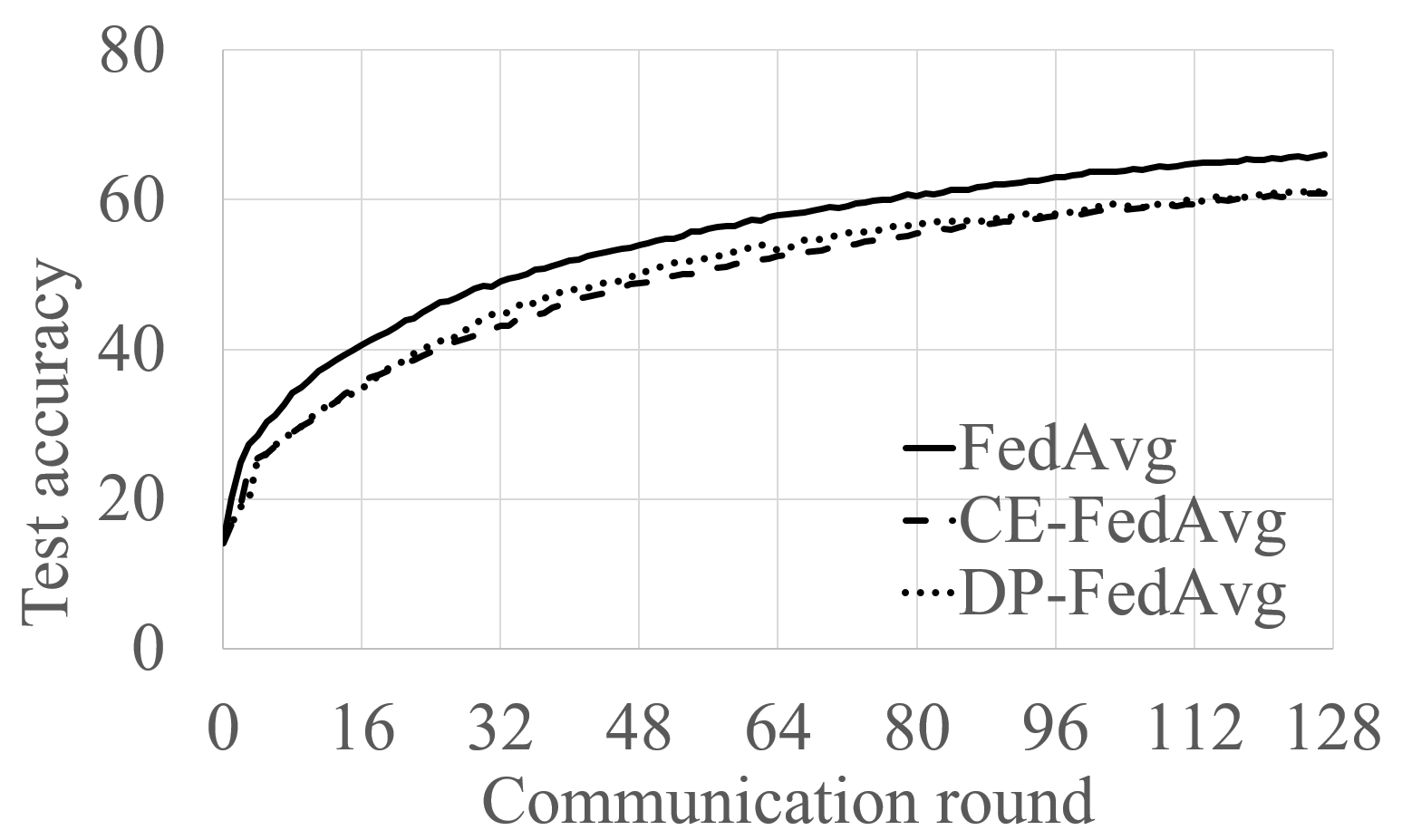}
         \caption{AlexNet, $\epsilon = 1.5$}
     \end{subfigure}
     \hfill
     \begin{subfigure}[b]{0.30\textwidth}
         \centering
         \includegraphics[width=\textwidth]{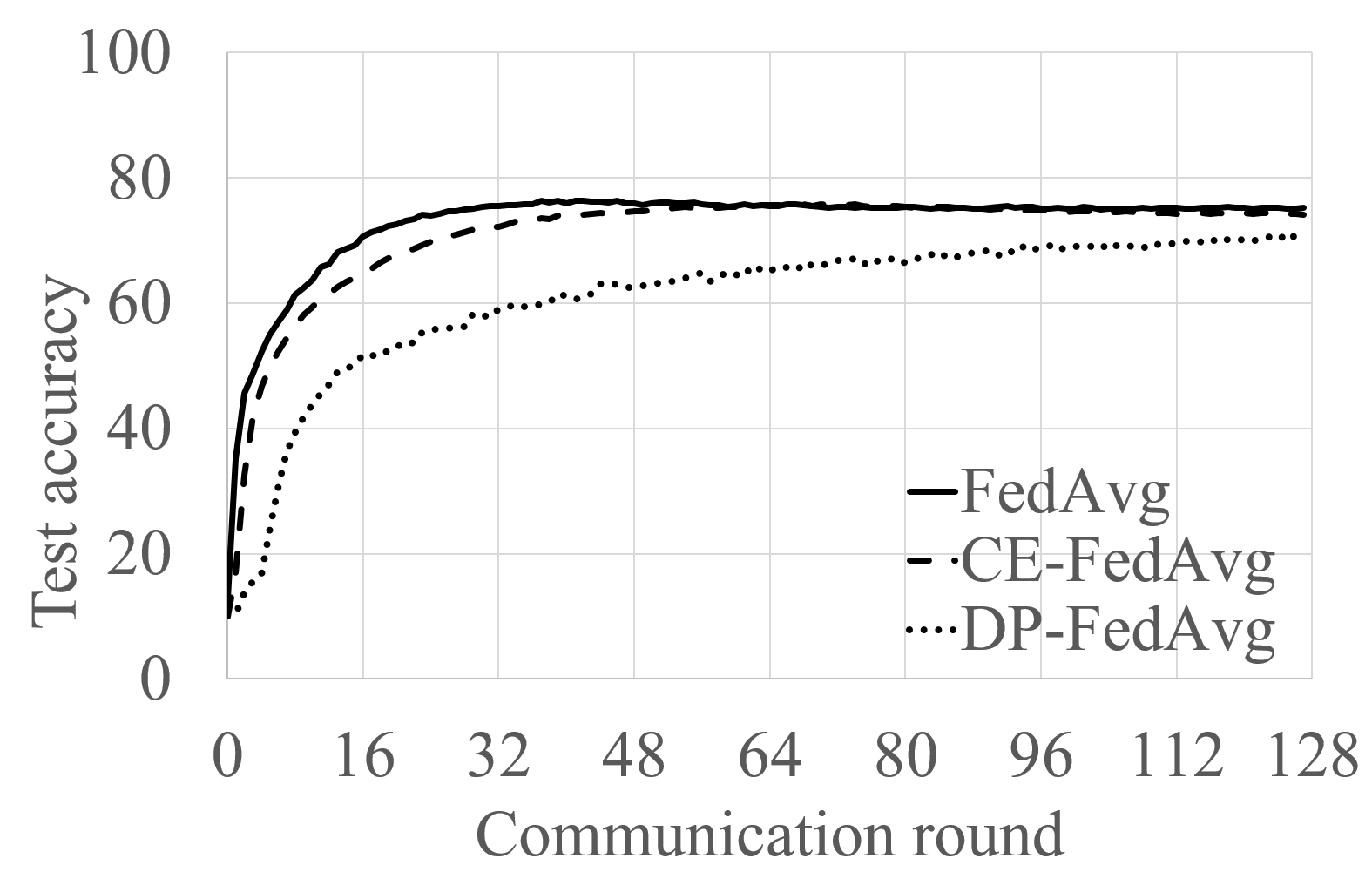}
         \caption{ResNet-18, $\epsilon = 1.5$}
     \end{subfigure}
    \caption{\footnotesize The test accuracy of FedAvg, CE-FedAvg and DP-FedAvg on different models on Cifar-10. The privacy budgets for MLP, AlexNet and ResNet are $\epsilon = 1.5$.}
    \label{fig:test_acc_dp2}
    \vspace{-0.5cm}
\end{figure}

\noindent{\bf Cifar-10 dataset.} The dataset we use is the Cifar-10 dataset, which has 50K training samples and 10K testing samples. We distribute the data in the IID way described in Section II and each client has 500 samples. We conduct experiments on a 2-layer MLP with one hidden layer, AlexNet and ResNet-18. The results are listed in Table~\ref{tab:performance2} and \figref{fig:test_acc_dp2}.

{\begin{table}[b!]
	\centering
	\begin{tabular}{l|rrrrr}
		\hline
			Model&  \# Parameters & \# Layers & Accuracy (\%)& Clipping (\% drop) & DP (\% drop)\\
			\hline
			MLP& 616K & 2 & 51.90 & 7.39 & 0.90\\
			AlexNet & 3.3M & 7 & 66.01 & 4.83 & -0.18\\
			ResNet-18 & 11.1M & 18 & 76.36 & 0.53 & 5.15 \\
			\hline
	\end{tabular}
	\caption{\footnotesize The accuracy drop between a) FedAvg and CE-FedAvg and b) CE-FedAvg and DP-FedAvg. The clipping threshold is $0.5$ of the average magnitude and privacy budget $\epsilon = 1.5$ for MLP, AlexNet and ResNet-18.}
	\label{tab:performance2}
\vspace{-0.5cm}
\end{table}}

\noindent{\bf Discussion.} Let us discuss the relation between our empirical observations and the theoretical results. 

{\bf 1)} It appears that when the underlying machine learning model is {\it structured} (e.g., many layers, has convolution layers, skip connections, etc), the update difference of FedAvg becomes {\it concentrated}, yielding a better clipping performance (as suggested by the terms related to clipping in Theorem \ref{thm: fedavg_conv});

{\bf 2)} When the model has too many parameters and/or layers, they are sensitive to privacy noise. This is reasonable since the error term caused by privacy noise in Theorem \ref{thm: fedavg_conv} is linearly dependent on the size of the model $d$ and the square of the Lipschitz constant $L$ (note, that $\eta_\ell\propto 1/L$). From~\cite[Corollary 3.3]{herrera2020estimating}, we know that $L$ increases exponentially with the number of layers. Therefore, larger and deeper models are potentially more sensitive to privacy noise.

{\bf 3)} We conjecture that, to ensure good performance of DP-FedAvg, we need to pick a neural network that is structured enough, while not having too many variables and too many number of layers.

\section{Conclusion}
This work provides empirical and theoretical understanding about clipping operation in FL. We show how to properly combine the clipping operation with existing FL algorithms to achieve the desirable trade-off between convergence and differential privacy guarantees.   Numerical results corroborate our theory, and suggest that the distribution of the clients' updates is a key factor that affects the performance of the clipping-enabled FL algorithm.  

\bibliography{References,ref-bi}
\bibliographystyle{IEEEbib}

\newpage
\appendix

\section{Appendix}
\subsection{Proof of Theorem \ref{thm: fedavg_conv}}
By Lipschitz smoothness, we have 
\begin{align}\label{eq:lip}
    f(x_{t+1}) \leq f(x_t) + \langle \nabla f(x_t), x_{t+1} - x_t \rangle + \frac{L}{2} \|x_{t+1} - x_t\|^2.
\end{align}
Before we proceed, we define following quantities to simplify notation:
\begin{align}
&\alpha_i^t := \frac{c}{\max(c,\eta_l \|\sum_{q=0}^{Q-1} g_i^{t,q} \|)}, \quad 
\tilde \alpha_i^t := \frac{c}{\max(c,\eta_l  \| \mathbb E[ \sum_{q=0}^{Q-1} g_i^{t,q} ]\|)},  \quad 
\overline \alpha^t := \frac{1}{N} \sum_{i=1}^N \tilde \alpha_i^t, \nonumber \\
&\Delta_i^t := -\eta_l \sum_{q=0}^{Q-1} g_i^{t,q} \cdot \alpha_i^t , \quad
\tilde \Delta_i^t:= -\eta_l \sum_{q=0}^{Q-1} g_i^{t,q} \cdot \tilde \alpha_i^t, \nonumber \\
&\overline \Delta_i^t := -\eta_l \sum_{q=0}^{Q-1} g_i^{t,q} \cdot \overline \alpha^t  ,\quad \breve  \Delta_i^t := -\eta_l \sum_{q=0}^{Q-1} \nabla f_i(x_i^{t,q}) \cdot \overline \alpha^t  \quad P := \abs{\cP_t},
\end{align}
where the expectation in $\tilde \alpha_i^t$ is taken over all possible randomness.

By using the above definitions, the model difference between two consecutive iterations can be expressed as:  $$x_{t+1} - x_t = \eta_g \frac{1}{P} \sum_{i\in\cP_t} (\Delta_i^t + z_i^t),$$ 
with $z_i^t \sim \mathcal N (0, \sigma^2 I)$.  Using the above expressions, and take an conditional expectation of \eqref{eq:lip} (conditioned on $x_t$), we obtain:
\begin{align}\label{eq: lip}
    \mathbb E [f(x_{t+1})] &\leq f(x_t)  + \eta_g {\left\langle \nabla f(x_t), \mathbb E\left[\frac{1}{P} \sum_{i\in\cP_t} \Delta_i^t + z_i^t\right] \right\rangle} + \frac{L}{2} \eta_g^2 \mathbb E \left [\left \|\frac{1}{P} \sum_{i\in\cP_t} \Delta_i^t + z_i^t\right\|^2\right]\nonumber\\
    &= f(x_t)  + \eta_g {\left\langle \nabla f(x_t), \mathbb E\left[\frac{1}{P} \sum_{i\in\cP_t} \Delta_i^t\right] \right\rangle} + \frac{L}{2} \eta_g^2 \mathbb E \left [\left \|\frac{1}{P} \sum_{i\in\cP_t} \Delta_i^t \right\|^2\right] + \frac{L}{2} \eta_g^2 \frac{1}{P} \sigma^2 d,
\end{align}
where $d$ in the last expression represents dimension of $x_t$; { in the last equation we use the fact that $z^t_i$ is zero mean.}

Next, we will analyze the bias caused by clipping, through analyzing the first order term in  \eqref{eq: lip}. Towards this end, we have the following series of relations: 
\begin{align}\label{{eq: bias_split}}
   & \left\langle \nabla f(x_t), \mathbb E\left[\frac{1}{P} \sum_{i\in\cP_t} \Delta_i^t\right] \right\rangle \nonumber\\
   & \stackrel{(i)}= \left\langle \nabla f(x_t), \mathbb E\left[\frac{1}{P}\E_i [\sum_{i\in\cP_t} \Delta_i^t]\right] \right\rangle =  \left\langle \nabla f(x_t), \frac{1}{P}P\mathbb E\left[\frac{1}{N} \sum_{i=1}^N \Delta_i^t\right] \right\rangle\nonumber \\
& = \left\langle \nabla f(x_t), \mathbb E\left[\frac{1}{N} \sum_{i=1}^N \Delta_i^t - \tilde \Delta_i^t \right] \right\rangle  + \left\langle \nabla f(x_t), \mathbb E\left[\frac{1}{N} \sum_{i=1}^N \tilde \Delta_i^t - \overline \Delta_i^t\right] \right\rangle  \nonumber \\
& +  \left\langle \nabla f(x_t), \mathbb E\left[\frac{1}{N} \sum_{i=1}^N \overline \Delta_i^t\right] \right\rangle 
\end{align}
{ where $(i)$ we takes expectation on the randomness of the client sampling, i.e., $\E_i \Delta_i^t = \frac{1}{N}\sum^N_{i=1}\Delta_i^t$}. 
The first two terms of RHS of the above equality can be viewed as bias caused by clipping. The first order predicted descent can be analyzed from the last term by completing the square:
\begin{align}
   & \left\langle \nabla f(x_t), \mathbb E\left[\frac{1}{N} \sum_{i=1}^N \overline \Delta_i^t\right] \right\rangle  \nonumber \\
   \stackrel{(i)}= & \mathbb  E\left[ \left\langle \nabla f(x_t), \frac{1}{N} \sum_{i=1}^N \breve \Delta_i^t\right] \right\rangle  \nonumber \\
   \stackrel{(ii)}=& \frac{-\eta_l \overline \alpha^t Q}{2} \|\nabla f(x_t)\|^2 -  \frac{\eta_l\overline \alpha^t}{2Q} \mathbb E \left[ \left\| \frac{1}{\eta_l N \overline \alpha^t} \sum_{i=1}^N \breve \Delta_i^t\right\|^2\right] \nonumber \\
  &+ \frac{\eta_l \overline \alpha^t}{2}  \underbrace{ \mathbb E \left[\left\|\sqrt{Q} \nabla f(x_t) - \frac{1}{\sqrt{Q}} \frac{1}{\eta_l N \overline \alpha^t}  \sum_{i=1}^N \breve \Delta_i^t\right \|^2 \right]}_{A_1},
\end{align}
{ where $(i)$ comes from $\E \overline \Delta_i^t = \breve \Delta_i^t$, $(ii)$ is because $\lin{a,b} = -\frac{1}{2}\norm{a}^2 - \frac{1}{2}\norm{b}^2 +\frac{1}{2}\norm{a-b}^2$ holds true for any vector $a,b$.}

We further upper bound $A_1$ as
\begin{align}\label{eq: A1}
    A_1  
    = & Q  \mathbb E \left[\left \| \nabla f(x_t) - \frac{1}{{Q}N} \sum_{i=1}^N \sum_{q=0}^{Q-1} \nabla f_i(x_i^{t,q}) \right \|^2\right] \nonumber \\
    =  & Q  \mathbb E \left[\left \|   \frac{1}{{Q}N} \sum_{i=1}^N \sum_{q=0}^{Q-1} \nabla f_i(x^t) -  \nabla f_i(x_i^{t,q})  \right \|^2 \right] \nonumber \\
    \leq & \frac{1}{N} \sum_{i=1}^N \sum_{q=0}^{Q-1} \mathbb E [ \| \nabla f_i(x^t) - \nabla f_i(x_i^{t,q}) \|^2 ]  \nonumber \\
    \leq & \frac{1}{N} \sum_{i=1}^N \sum_{q=0}^{Q-1} L^2 \mathbb E [ \|  x^t - x_i^{t,q} \|^2 ]  \nonumber \\
    \leq & L^2 5 Q^2 \eta_l^2 (\sigma_l^2 + 6Q \sigma_g^2) + L^2 30 Q^3 \eta_l^2 \|\nabla f(x_t)\|^2
\end{align}
{ where the first inequality comes from Jensen's inequality, the second inequality comes from $L$-smoothness} and the last inequality is due to { \cite[Lemma 3]{reddi2020adaptive}, that the following inequality holds for any $q \in \{0,\dots, Q-1\}$}
\begin{equation*}
    \frac{1}{N}\sum^N_{i=1} \E \left[ \norm{x^t - x_i^{t,q}}^2 \right] \leq 5Q\eta_l^2(\sigma_l^2+6Q\sigma_g^2) +30Q^2\eta_l^2\norm{\nabla f(x_t)}^2.
\end{equation*}


Now we turn to upper bounding the second order term in \eqref{eq: lip}, as follows
\begin{align}\label{eq: second_term}
     & \mathbb E \left [\left \|\frac{1}{P} \sum_{i\in\cP_t} \Delta_i^t \right\|^2\right] \nonumber \\ 
     \leq & 3 \mathbb E \left [\left \|\frac{1}{P} \sum_{i\in\cP_t} \Delta_i^t - \tilde \Delta_i^t \right\|^2\right] + 3 \mathbb E \left [\left \|\frac{1}{P} \sum_{i\in\cP_t} \tilde \Delta_i^t - \overline \Delta_i^t  \right\|^2\right] + 3 \mathbb E \left [\left \|\frac{1}{P} \sum_{i\in\cP_t}  \overline \Delta_i^t  \right\|^2\right] .
\end{align}
We can bound the expectation in the last term of \eqref{eq: second_term} as follows: 
\begin{align}\label{eq: split_variance}
    &\mathbb E \left [\left \|\frac{1}{P} \sum_{i\in\cP_t} \overline \Delta_i^t \right\|^2\right]  \nonumber \\
    = & \mathbb E \left [\left \|\frac{1}{P} \sum_{i\in\cP_t} \left(\eta_l \sum_{q=0}^{Q-1} g_i^{t,q} \cdot \overline \alpha^t\right) \right\|^2\right] \nonumber \\
    \leq & \eta_l^2\mathbb E \left [ 2 \left \|\frac{1}{P} \sum_{i\in\cP_t} \sum_{q=0}^{Q-1} \nabla f(x_i^{t,q}) \cdot \overline \alpha^t \right\|^2 + 2 \left \|\frac{1}{P} \sum_{i\in\cP_t} \sum_{q=0}^{Q-1} (\nabla f(x_i^{t,q}) -g_i^{t,q}) \cdot \overline \alpha^t \right\|^2\right] \nonumber \\
    \leq & 2 \mathbb E \left [\left \|\frac{1}{P} \sum_{i\in\cP_t} \breve \Delta_i^t \right\|^2\right] + \frac{2}{P}\eta_l^2\overline \alpha^2 Q \sigma_l^2
\end{align}
{  where the last inequality is because the assumption that $\mathbb E [\|g_i^{t,q} - \nabla f_i(x_i^{t,q})\|^2] \leq \sigma_l^2$}. 
Let us further bound the expectation in the first term of \eqref{eq: split_variance} as:
\begin{equation}\label{eq: subsample}
    \begin{aligned}
        \E \left [\norm{\frac{1}{P} \sum_{i\in\cP_t} \breve \Delta_i^t }^2\right] &= \frac{1}{P^2}\E\left [\norm{ \sum_{i\in\cP_t} \breve \Delta_i^t }^2\right]\\
        & \stackrel{(i)}= \frac{1}{P^2}\E \left [\E_i\sum_{i\in\cP_t} \norm{ \breve \Delta_i^t }^2 + \E_{i,j} \sum_{i\neq j\in\cP_t}\lin{\breve \Delta_i^t, \breve \Delta_j^t}\right]\\
        & \stackrel{(ii)}= \frac{1}{P^2}\E \left [\frac{P}{N}\sum^N_{i = 1} \norm{ \breve \Delta_i^t }^2 + P(P-1)\lin{\E_i\breve \Delta_i^t, \E_j\breve \Delta_j^t}\right]\\
        & = \frac{1}{P^2}\E \left [\frac{P}{N}\sum^N_{i = 1} \norm{ \breve \Delta_i^t }^2 + P(P-1)\norm{\frac{1}{N}\sum^N_{i=1}\breve \Delta_i^t}^2\right],
    \end{aligned}
\end{equation}
where in (i) we expand the square and take expectation on the randomness of client sampling, and (ii) is due to independent sampling the clients {\it with} replacement so that $\E_{i,j}\lin{\Delta_i^t,\Delta_j^t} = \lin{\E_i \Delta_i^t, \E_j \Delta_j^t}.$

{ Additionally, note we have:}  
\begin{equation}\label{eq: subsample_1}
    \begin{aligned}
        \E \sum^N_{i = 1} \norm{ \breve \Delta_i^t }^2 & \stackrel{(i)}=\E \sum^N_{i = 1} \eta_l^2(\overline\alpha^t)^2\norm{\sum^{Q-1}_{q=0} \nabla f_i(x^t) + \nabla f_i(x_i^{t,q})  - \nabla f_i(x^t)}^2\\
        & \stackrel{(ii)}\leq 2\eta_l^2\overline\alpha^t\sum^N_{i = 1}\left(Q^2\sum^{Q-1}_{q=0}  \E \norm{\nabla f_i(x^t) + \nabla f_i(x_i^{t,q})}^2+  Q^2\sum^{Q-1}_{q=0}\norm{\nabla f_i(x^t)}^2 \right)\\
        & \stackrel{(iii)}\leq 2\eta_l^2\overline \alpha^tN\bigg(L^2 5 Q^2 \eta_l^2 (\sigma_l^2 + 6Q \sigma_g^2) + L^2 30 Q^3 \eta_l^2 \|\nabla f(x_t)\|^2 + 2Q^3\|\nabla f(x_t)\|^2+ 2Q^3\sigma^2_g\bigg)\\
        & = 10N\eta_l^4\overline\alpha^tL^2Q^2\sigma_l^2 + 4N\eta_l^2\overline \alpha^tQ^3(15L^2\eta_l^2+1) (\|\nabla f(x_t)\|^2+\sigma^2_g).
    \end{aligned}
\end{equation}
{ where $(i)$ comes from the definition of $\breve \Delta_i^t$; $(ii)$ comes from the fact that $\norm{a+b}^2\leq 2(\norm{a}^2 + \norm{b}^2$); in $(iii)$ we apply \eqref{eq: A1} to the first term and bound the second term by the assumption that $\|\nabla f_i(x) - \nabla f(x)\|^2 \leq \sigma_g^2$.}

Combining \eqref{eq: lip}-\eqref{eq: subsample_1}, we have
\begin{align}
        \mathbb E [f(x_{t+1})] \leq & f(x_t)  - \frac{\eta_g\eta_l \overline \alpha^t Q}{2} \|\nabla f(x_t)\|^2 - \frac{\eta_g\eta_l\overline \alpha^t}{2Q} \mathbb E \left[ \left\| \frac{1}{\eta_l N \overline \alpha^t} \sum_{i=1}^N \breve \Delta_i^t\right\|^2\right] \nonumber \\
  &+ \frac{\eta_g\eta_l \overline \alpha^t}{2} (5L^2  Q^2 \eta_l^2 (\sigma_l^2 + 6Q \sigma_g^2) + 30L^2  Q^3 \eta_l^2 \|\nabla f(x_t)\|^2) \nonumber \\
        &+ \eta_g \left\langle \nabla f(x_t), \mathbb E\left[\frac{1}{N} \sum_{i=1}^N \Delta_i^t - \tilde \Delta_i^t \right] \right\rangle  + \eta_g \left\langle \nabla f(x_t), \mathbb E\left[\frac{1}{N} \sum_{i=1}^N \tilde \Delta_i^t - \overline \Delta_i^t\right] \right\rangle \nonumber \\
        & + \frac{3L\eta_g^2(P-1)}{P} \mathbb E \left [\left \|\frac{1}{N} \sum^N_{i = 1} \breve \Delta_i^t \right\|^2\right]  +  \frac{3L}{P}\eta_g^2\eta_l^2(\overline \alpha^t)^2 Q \sigma_l^2 + \frac{L}{2} \eta_g^2 \frac{1}{P} \sigma^2 d   \nonumber \\
        & + \frac{30}{P}\eta_l^4\eta_g^2\overline\alpha^tL^2Q^2\sigma_l^2 + \frac{12}{P}\eta_l^2\eta_g^2\overline \alpha^tQ^3(15L^2\eta_l^2 + 1) (\|\nabla f(x_t)\|^2 + \sigma^2_g) \nonumber\\
        & + \frac{3L}{2} \eta_g^2 \mathbb E \left [\left \|\frac{1}{P} \sum_{i\in\cP_t} \Delta_i^t - \tilde \Delta_i^t  \right\|^2\right]  + \frac{3L}{2} \eta_g^2 \mathbb E \left [\left \|\frac{1}{P} \sum_{i\in\cP_t} \tilde\Delta_i^t -  \overline\Delta_i^t  \right\|^2\right]  
\end{align}
When $\eta_g\eta_l \leq \min\{\frac{\sqrt{P}}{\sqrt{48Q}Q},\frac{P}{6QL(P-1)}\}$ and $\eta_l \leq \frac{1}{\sqrt{60}QL}$, the above inequality simplifies to  
\begin{align}
    \mathbb E [f(x_{t+1})] 
    \leq & f(x_t)  - \frac{\eta_g\eta_l \overline \alpha^t Q}{4} \|\nabla f(x_t)\|^2  \nonumber \\
  &+ \frac{5\eta_g\eta_l^3 \overline \alpha^t}{2}(1+\frac{12\eta_l\eta_g}{P})L^2  Q^2  (\sigma_l^2 + 6Q \sigma_g^2) \nonumber \\
        &+ \eta_g \left\langle \nabla f(x_t), \mathbb E\left[\frac{1}{N} \sum_{i=1}^N \Delta_i^t - \tilde \Delta_i^t \right] \right\rangle  + \eta_g \left\langle \nabla f(x_t), \mathbb E\left[\frac{1}{N} \sum_{i=1}^N \tilde \Delta_i^t - \overline \Delta_i^t\right] \right\rangle \nonumber \\
        & + \frac{3L}{N}\eta_g^2\eta_l^2(\overline \alpha^t)^2 Q \sigma_l^2 + \frac{L}{2} \eta_g^2 \frac{1}{P} \sigma^2 d   \nonumber \\
        & + \frac{3L}{2} \eta_g^2 \mathbb E \left [\left \|\frac{1}{P} \sum_{i\in\cP_t} \Delta_i^t - \tilde \Delta_i^t  \right\|^2\right]  + \frac{3L}{2} \eta_g^2 \mathbb E \left [\left \|\frac{1}{P} \sum_{i\in\cP_t} \tilde\Delta_i^t -  \overline\Delta_i^t  \right\|^2\right]  
\end{align}
Sum over $t$ from $1$ to $T$, divide both sides by $T\eta_g\eta_lQ/4$, and rearrange, we have
\begin{align}
      &\frac{1}{T} \sum_{t=1}^T \mathbb E[ \overline \alpha^t\|\nabla f(x_t)\|^2] \nonumber \\
    \leq & \frac{4}{T\eta_g\eta_lQ}(\mathbb E [f(x_{1})] -  \mathbb E [f(x_{T+1})])      \nonumber \\
  &+ {10\eta_l^2 } L^2  Q (1+\frac{12\eta_l\eta_g}{P}) (\sigma_l^2 + 6Q \sigma_g^2) \frac{1}{T}\sum_{t=1}^T \overline \alpha^t + \frac{12L}{P}\eta_g\eta_l  \sigma_l^2 \frac{1}{T}\sum_{t=1}^T (\overline \alpha^t)^2+  2L\frac{ \eta_g}{\eta_lQP} d \sigma^2    \nonumber \\
        &+ \frac{1}{T}\sum_{t=1}^T \frac{4}{\eta_lQ} \mathbb E\left[\left\langle \nabla f(x_t), \mathbb E\left[\frac{1}{N} \sum_{i=1}^N \Delta_i^t - \tilde \Delta_i^t \right] \right\rangle  + \left\langle \nabla f(x_t), \mathbb E\left[\frac{1}{N} \sum_{i=1}^N \tilde \Delta_i^t - \overline \Delta_i^t\right] \right\rangle \right ] \nonumber \\
        & + \frac{6L}{\eta_lQ} \eta_g \frac{1}{T} \sum_{t=1}^T \mathbb E \left [\left \|\frac{1}{P} \sum_{i\in\cP_t} \Delta_i^t - \tilde \Delta_i^t  \right\|^2\right]  + \frac{6L}{\eta_lQ} \eta_g \frac{1}{T} \sum_{t=1}^T \mathbb E \left [\left \|\frac{1}{P} \sum_{i\in\cP_t} \tilde\Delta_i^t -  \overline\Delta_i^t  \right\|^2\right].  
\end{align}
Upper-bounding the last four terms using $\|g_i^{t,q}\| \leq G$ yields the desired result. \qed

\subsection{Additional Numerical Experiments}\label{app:exp}
In this part, we provide additional numerical results. 
\subsubsection{Update Distributions}

\begin{figure}[t!]
    \centering
     \begin{subfigure}[b]{0.21\textwidth}
         \centering         
		 \includegraphics[width=\textwidth]{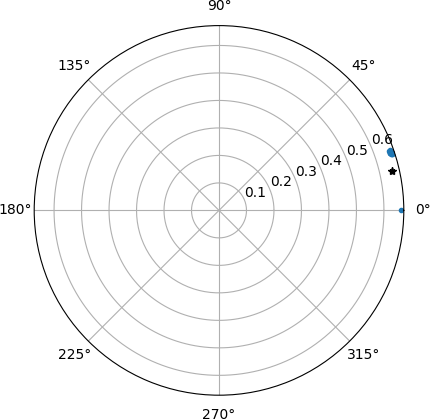}
         \caption{IID, $t = 0$}
     \end{subfigure}
     \hfill
     \begin{subfigure}[b]{0.21\textwidth}
         \centering
         \includegraphics[width=\textwidth]{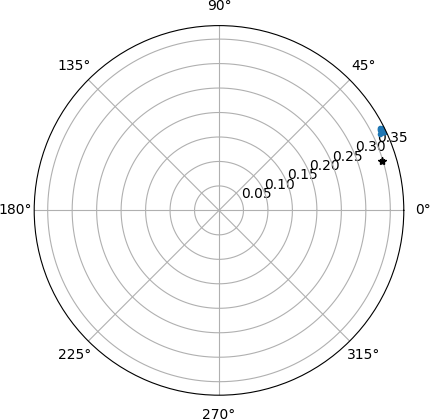}
         \caption{IID, $t = 2$}
     \end{subfigure}
     \begin{subfigure}[b]{0.21\textwidth}
         \centering
         \includegraphics[width=\textwidth]{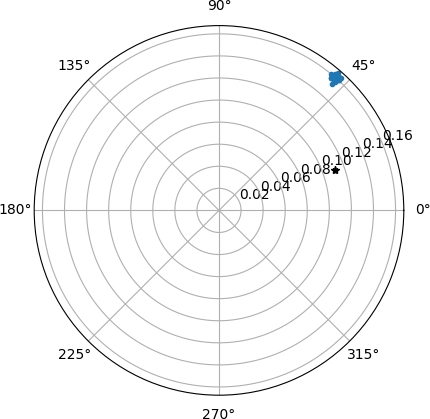}
         \caption{IID, $t = 8$}
     \end{subfigure}
     \hfill
     \begin{subfigure}[b]{0.21\textwidth}
         \centering
         \includegraphics[width=\textwidth]{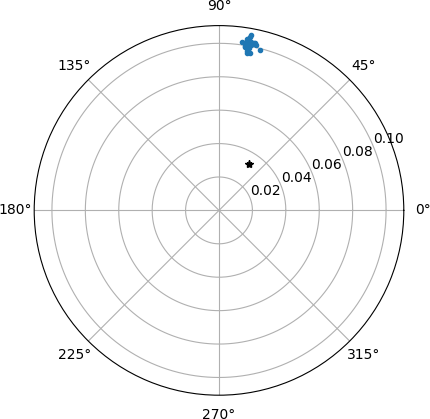}
         \caption{IID, $t = 64$}
     \end{subfigure}
     \begin{subfigure}[b]{0.21\textwidth}
         \centering         
		 \includegraphics[width=\textwidth]{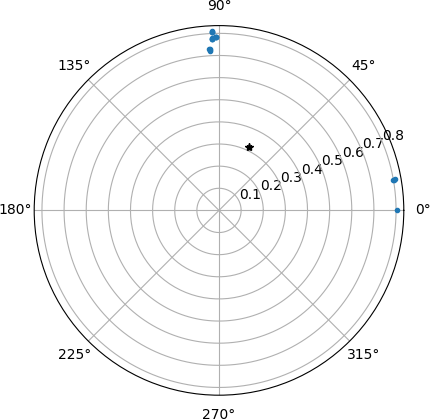}
         \caption{Non-IID, $t = 0$}
     \end{subfigure}
     \hfill
     \begin{subfigure}[b]{0.21\textwidth}
         \centering
         \includegraphics[width=\textwidth]{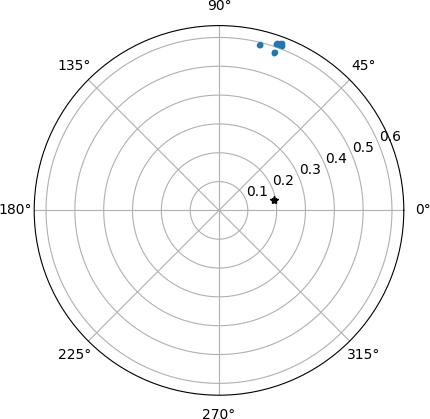}
         \caption{Non-IID, $t = 2$}
     \end{subfigure}
     \begin{subfigure}[b]{0.21\textwidth}
         \centering
         \includegraphics[width=\textwidth]{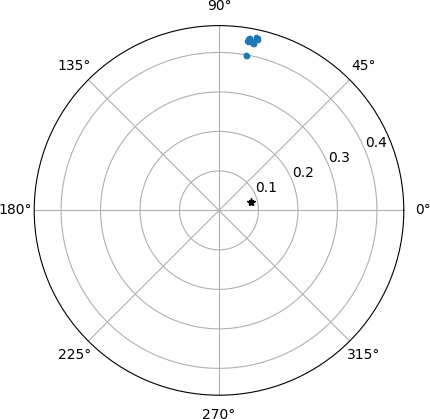}
         \caption{Non-IID, $t = 8$}
     \end{subfigure}
     \hfill
     \begin{subfigure}[b]{0.21\textwidth}
         \centering
         \includegraphics[width=\textwidth]{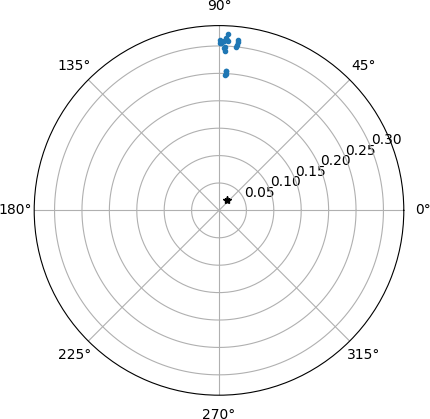}
         \caption{Non-IID, $t = 64$}
     \end{subfigure}
    \caption{\footnotesize The distribution of local updates for MLP on IID and Non-IID data at different communication rounds for EMNIST dataset. Each blue dot corresponds to the local update from one client. The black dot shows the magnitude and the cosine angle of global model update at iteration $t$.}
    \label{fig:app_MLP_emnist}
\end{figure}

\begin{figure}[b!]
    \centering
     \begin{subfigure}[b]{0.21\textwidth}
         \centering         
		 \includegraphics[width=\textwidth]{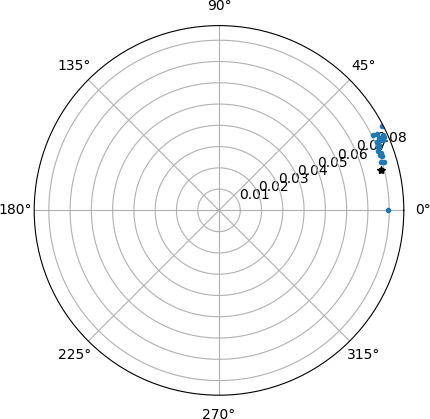}
         \caption{IID, $t = 0$}
     \end{subfigure}
     \hfill
     \begin{subfigure}[b]{0.21\textwidth}
         \centering
         \includegraphics[width=\textwidth]{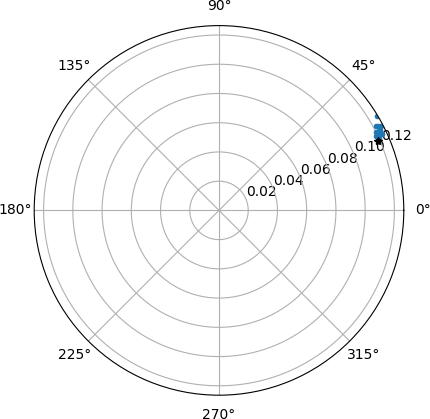}
         \caption{IID, $t = 2$}
     \end{subfigure}
     \begin{subfigure}[b]{0.21\textwidth}
         \centering
         \includegraphics[width=\textwidth]{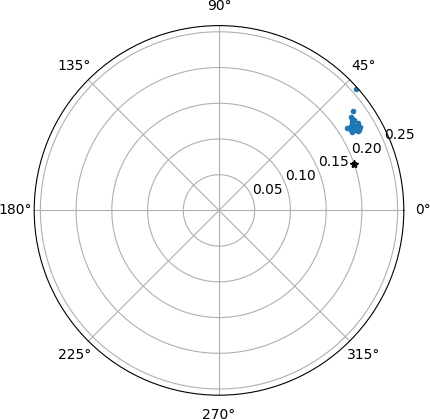}
         \caption{IID, $t = 8$}
     \end{subfigure}
     \hfill
     \begin{subfigure}[b]{0.21\textwidth}
         \centering
         \includegraphics[width=\textwidth]{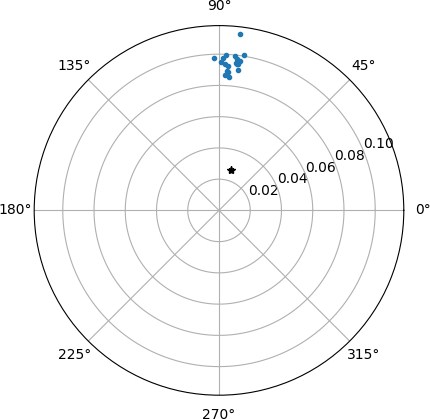}
         \caption{IID, $t = 64$}
     \end{subfigure}
     \begin{subfigure}[b]{0.21\textwidth}
         \centering         
		 \includegraphics[width=\textwidth]{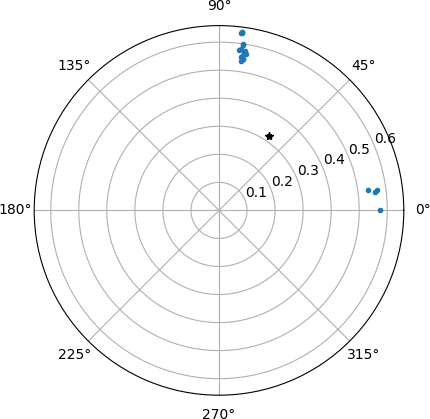}
         \caption{Non-IID, $t = 0$}
     \end{subfigure}
     \hfill
     \begin{subfigure}[b]{0.21\textwidth}
         \centering
         \includegraphics[width=\textwidth]{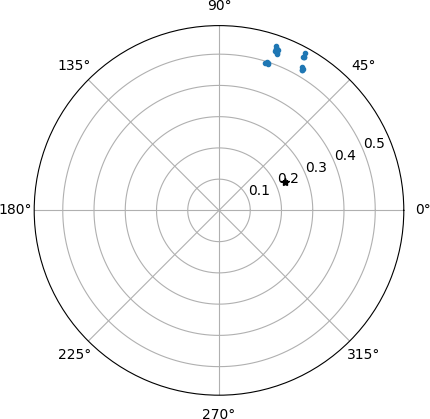}
         \caption{Non-IID, $t = 2$}
     \end{subfigure}
     \begin{subfigure}[b]{0.21\textwidth}
         \centering
         \includegraphics[width=\textwidth]{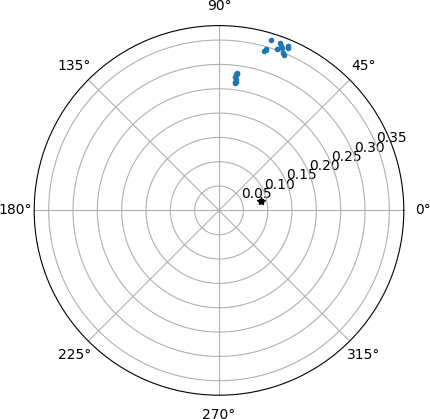}
         \caption{Non-IID, $t = 8$}
     \end{subfigure}
     \hfill
     \begin{subfigure}[b]{0.21\textwidth}
         \centering
         \includegraphics[width=\textwidth]{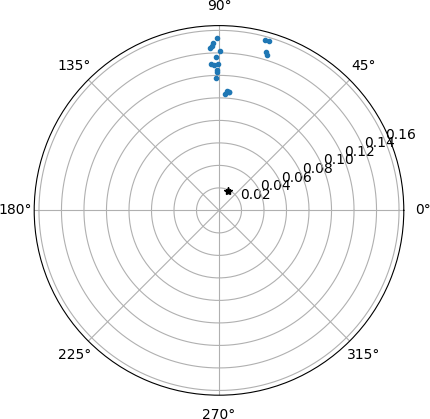}
         \caption{Non-IID, $t = 64$}
     \end{subfigure}
    \caption{\footnotesize The distribution of local updates for AlexNet on IID and Non-IID data at different communication rounds for EMNIST dataset. Each blue dot corresponds to the local update from one client. The black dot shows the magnitude and the cosine angle of global local model update at iteration $t$. }
    \label{fig:app_alex_emnist}
\end{figure}

\begin{figure}[t!]
    \centering
     \begin{subfigure}[b]{0.21\textwidth}
         \centering         
		 \includegraphics[width=\textwidth]{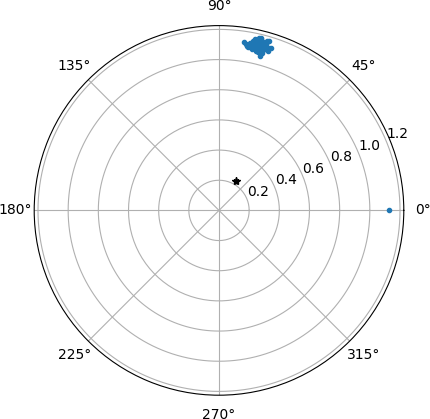}
         \caption{IID, $t = 0$}
     \end{subfigure}
     \hfill
     \begin{subfigure}[b]{0.21\textwidth}
         \centering
         \includegraphics[width=\textwidth]{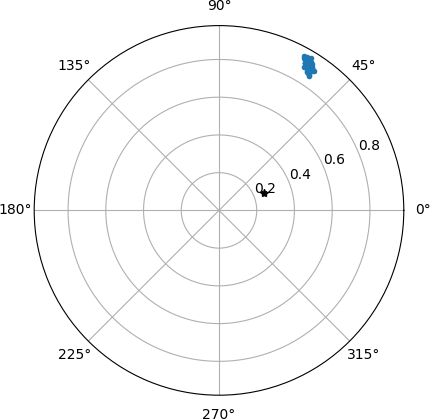}
         \caption{IID, $t = 2$}
     \end{subfigure}
     \begin{subfigure}[b]{0.21\textwidth}
         \centering
         \includegraphics[width=\textwidth]{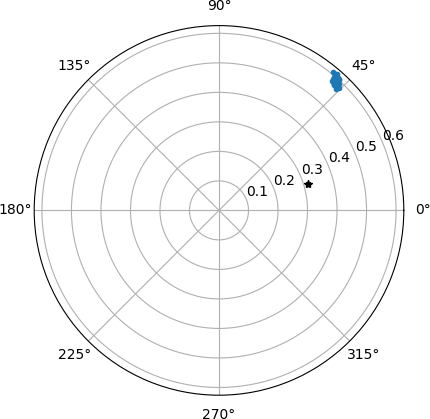}
         \caption{IID, $t = 8$}
     \end{subfigure}
     \hfill
     \begin{subfigure}[b]{0.21\textwidth}
         \centering
         \includegraphics[width=\textwidth]{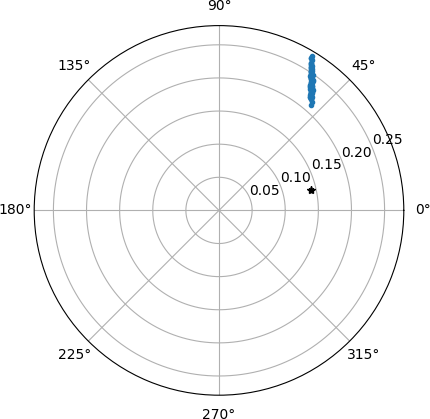}
         \caption{IID, $t = 32$}
     \end{subfigure}
     \begin{subfigure}[b]{0.21\textwidth}
         \centering         
		 \includegraphics[width=\textwidth]{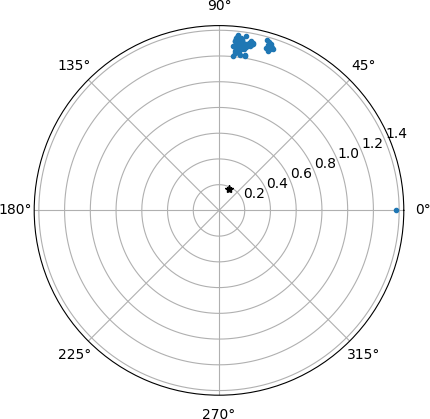}
         \caption{Non-IID, $t = 0$}
     \end{subfigure}
     \hfill
     \begin{subfigure}[b]{0.21\textwidth}
         \centering
         \includegraphics[width=\textwidth]{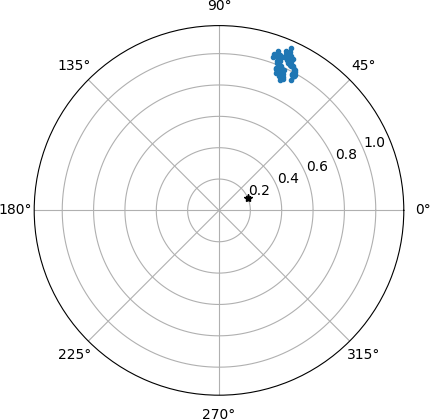}
         \caption{Non-IID, $t = 2$}
     \end{subfigure}
     \begin{subfigure}[b]{0.21\textwidth}
         \centering
         \includegraphics[width=\textwidth]{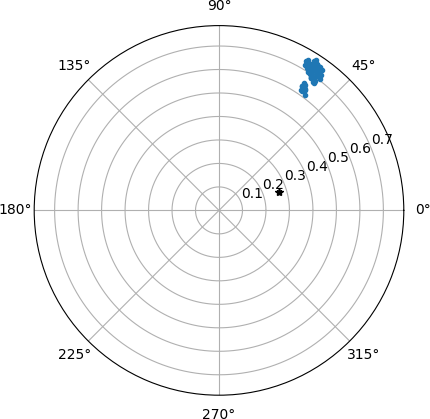}
         \caption{Non-IID, $t = 8$}
     \end{subfigure}
     \hfill
     \begin{subfigure}[b]{0.21\textwidth}
         \centering
         \includegraphics[width=\textwidth]{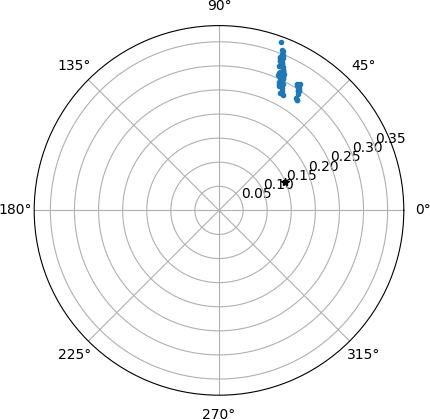}
         \caption{Non-IID, $t = 32$}
     \end{subfigure}
    \caption{\footnotesize The distribution of local updates for MobileNetV2 on IID and Non-IID data at different communication rounds for EMNIST dataset. Each blue dot corresponds to the local update from one client. The black dot shows the magnitude and the cosine angle of global local model update at iteration $t$.}
    \label{fig:app_mobile_emnist}
\end{figure}

\begin{figure}[b!]
    \centering
     \begin{subfigure}[b]{0.21\textwidth}
         \centering         
		 \includegraphics[width=\textwidth]{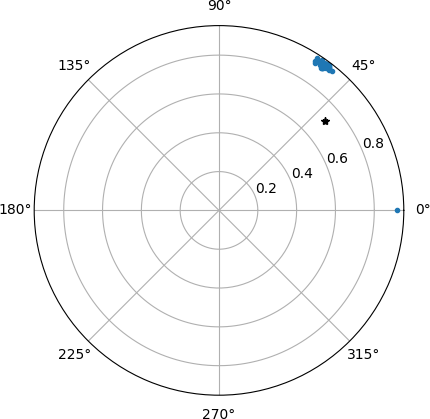}
         \caption{IID, $t = 0$}
     \end{subfigure}
     \hfill
     \begin{subfigure}[b]{0.21\textwidth}
         \centering
         \includegraphics[width=\textwidth]{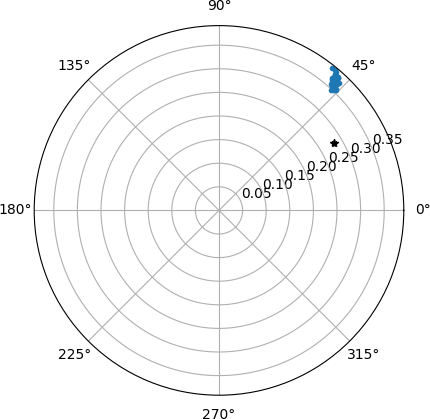}
         \caption{IID, $t = 2$}
     \end{subfigure}
     \begin{subfigure}[b]{0.21\textwidth}
         \centering
         \includegraphics[width=\textwidth]{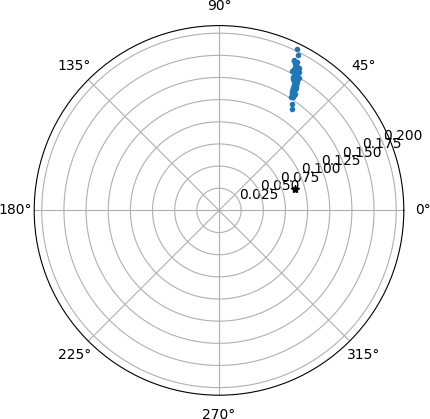}
         \caption{IID, $t = 8$}
     \end{subfigure}
     \hfill
     \begin{subfigure}[b]{0.21\textwidth}
         \centering
         \includegraphics[width=\textwidth]{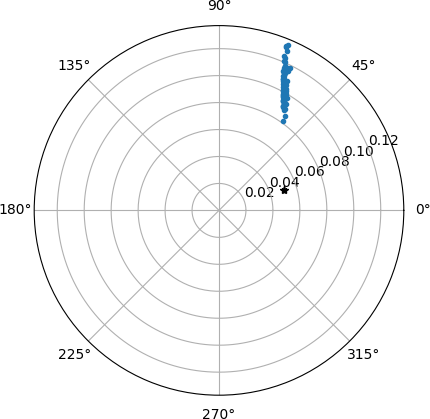}
         \caption{IID, $t = 32$}
     \end{subfigure}
     \begin{subfigure}[b]{0.21\textwidth}
         \centering         
		 \includegraphics[width=\textwidth]{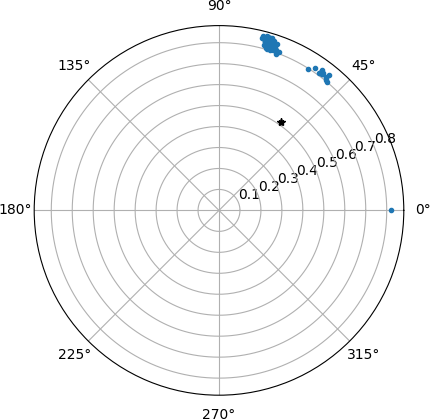}
         \caption{Non-IID, $t = 0$}
     \end{subfigure}
     \hfill
     \begin{subfigure}[b]{0.21\textwidth}
         \centering
         \includegraphics[width=\textwidth]{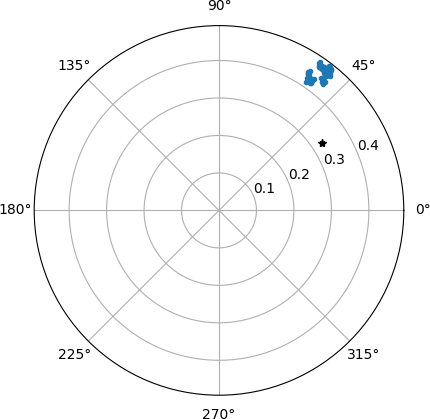}
         \caption{Non-IID, $t = 2$}
     \end{subfigure}
     \begin{subfigure}[b]{0.21\textwidth}
         \centering
         \includegraphics[width=\textwidth]{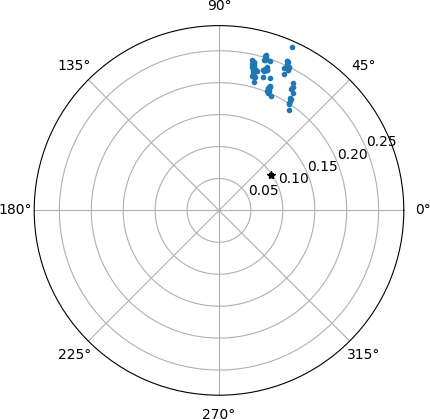}
         \caption{Non-IID, $t = 8$}
     \end{subfigure}
     \hfill
     \begin{subfigure}[b]{0.21\textwidth}
         \centering
         \includegraphics[width=\textwidth]{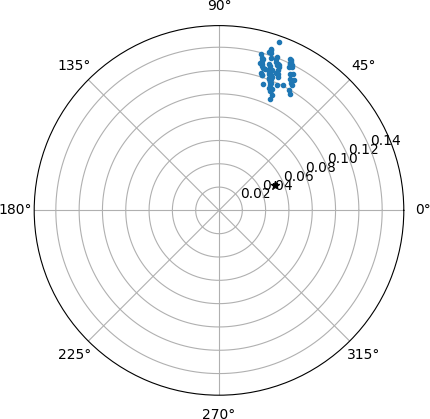}
         \caption{Non-IID, $t = 32$}
     \end{subfigure}
    \caption{\footnotesize The distribution of local updates for ResNet-18 on IID and Non-IID data at different communication rounds for EMNIST dataset. Each blue dot corresponds to the local update from one client. The black dot shows the magnitude and the cosine angle of global local model update at iteration $t$.}
    \label{fig:app_resnet_emnist}
\end{figure}

\begin{figure}[t!]
    \centering
     \begin{subfigure}[b]{0.21\textwidth}
         \centering         
		 \includegraphics[width=\textwidth]{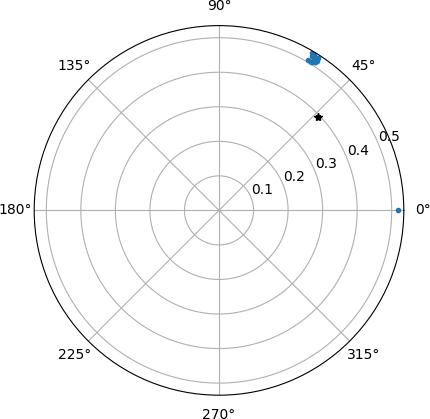}
         \caption{IID, $t = 0$}
     \end{subfigure}
     \hfill
     \begin{subfigure}[b]{0.21\textwidth}
         \centering
         \includegraphics[width=\textwidth]{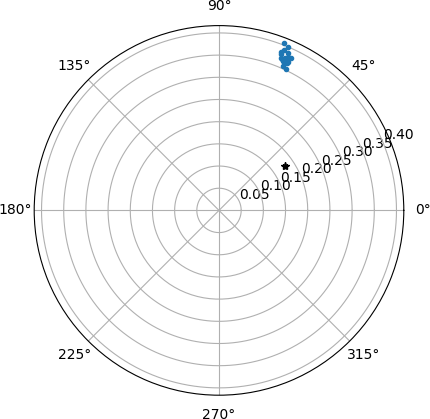}
         \caption{IID, $t = 2$}
     \end{subfigure}
     \begin{subfigure}[b]{0.21\textwidth}
         \centering
         \includegraphics[width=\textwidth]{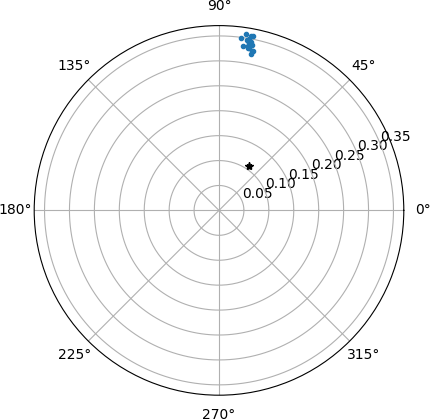}
         \caption{IID, $t = 8$}
     \end{subfigure}
     \hfill
     \begin{subfigure}[b]{0.21\textwidth}
         \centering
         \includegraphics[width=\textwidth]{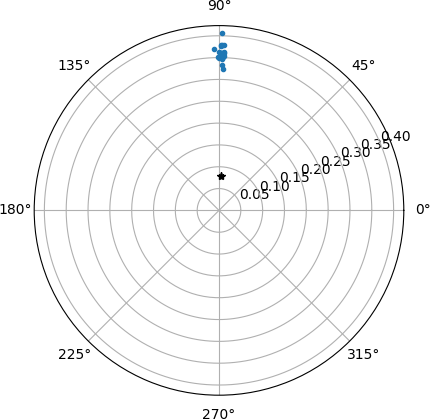}
         \caption{IID, $t = 64$}
     \end{subfigure}
     \begin{subfigure}[b]{0.21\textwidth}
         \centering         
		 \includegraphics[width=\textwidth]{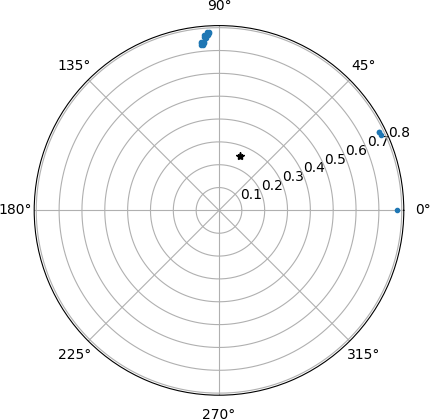}
         \caption{Non-IID, $t = 0$}
     \end{subfigure}
     \hfill
     \begin{subfigure}[b]{0.21\textwidth}
         \centering
         \includegraphics[width=\textwidth]{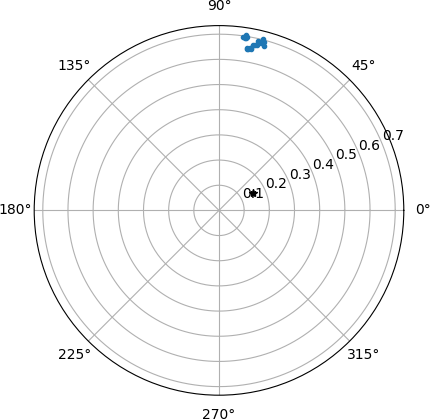}
         \caption{Non-IID, $t = 2$}
     \end{subfigure}
     \begin{subfigure}[b]{0.21\textwidth}
         \centering
         \includegraphics[width=\textwidth]{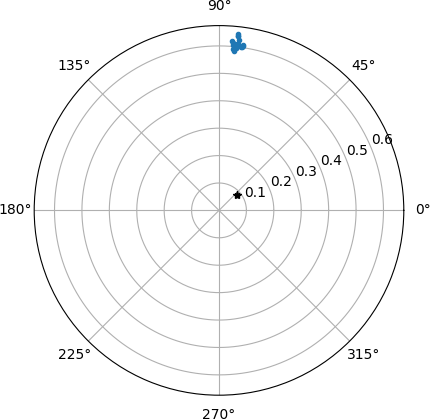}
         \caption{Non-IID, $t = 8$}
     \end{subfigure}
     \hfill
     \begin{subfigure}[b]{0.21\textwidth}
         \centering
         \includegraphics[width=\textwidth]{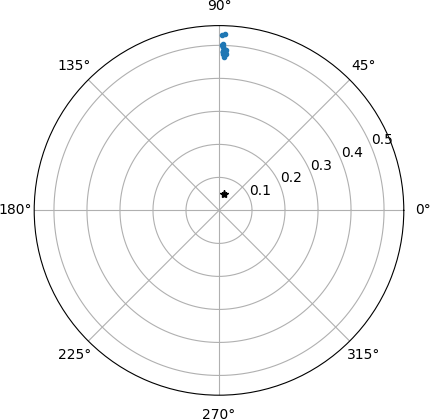}
         \caption{Non-IID, $t = 64$}
     \end{subfigure}
    \caption{\footnotesize The distribution of local updates for MLP on IID and Non-IID data at different communication rounds for Cifar-10 dataset. Each blue dot corresponds to the local update from one client. The black dot shows the magnitude and the cosine angle of global local model update at iteration $t$.}
    \label{fig:app_MLP_cifar}
\end{figure}

\begin{figure}[b!]
    \centering
     \begin{subfigure}[b]{0.21\textwidth}
         \centering         
		 \includegraphics[width=\textwidth]{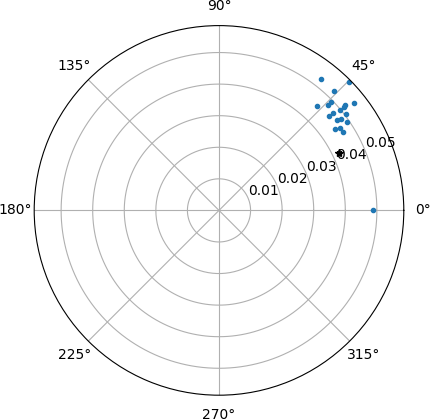}
         \caption{IID, $t = 0$}
     \end{subfigure}
     \hfill
     \begin{subfigure}[b]{0.21\textwidth}
         \centering
         \includegraphics[width=\textwidth]{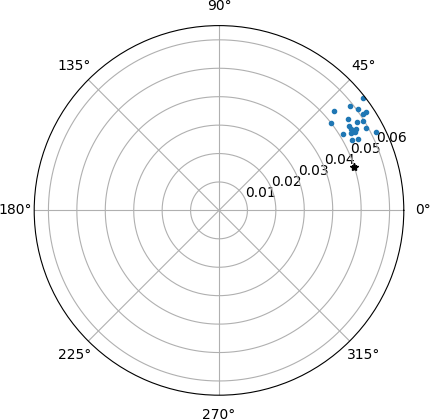}
         \caption{IID, $t = 2$}
     \end{subfigure}
     \begin{subfigure}[b]{0.21\textwidth}
         \centering
         \includegraphics[width=\textwidth]{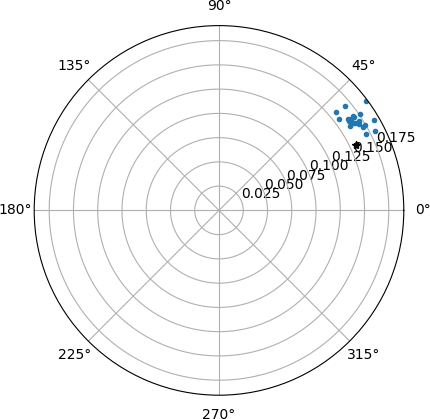}
         \caption{IID, $t = 8$}
     \end{subfigure}
     \hfill
     \begin{subfigure}[b]{0.21\textwidth}
         \centering
         \includegraphics[width=\textwidth]{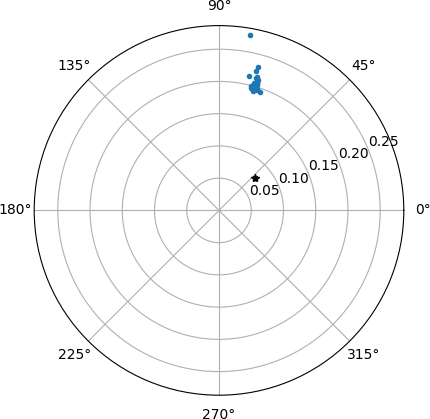}
         \caption{IID, $t = 64$}
     \end{subfigure}
     \begin{subfigure}[b]{0.21\textwidth}
         \centering         
		 \includegraphics[width=\textwidth]{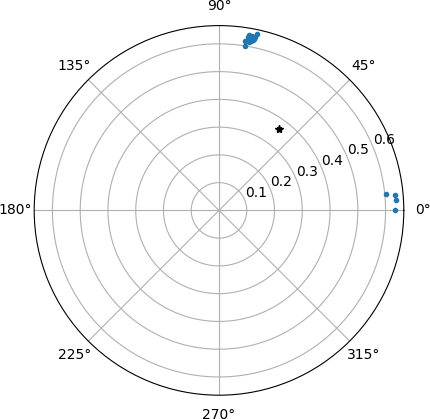}
         \caption{Non-IID, $t = 0$}
     \end{subfigure}
     \hfill
     \begin{subfigure}[b]{0.21\textwidth}
         \centering
         \includegraphics[width=\textwidth]{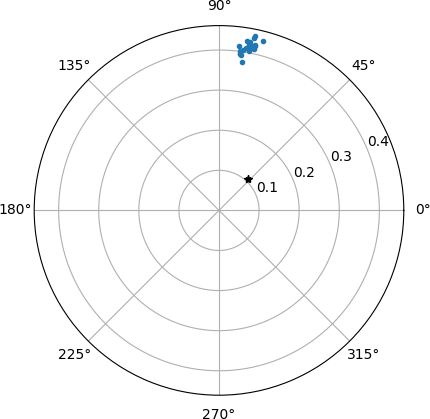}
         \caption{Non-IID, $t = 2$}
     \end{subfigure}
     \begin{subfigure}[b]{0.21\textwidth}
         \centering
         \includegraphics[width=\textwidth]{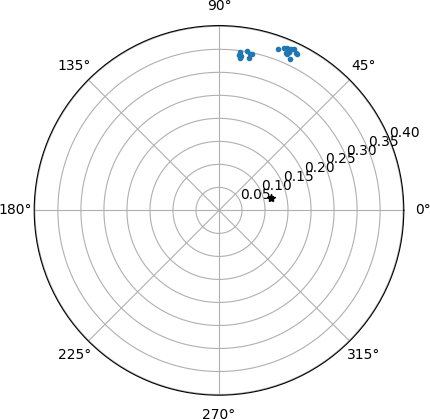}
         \caption{Non-IID, $t = 8$}
     \end{subfigure}
     \hfill
     \begin{subfigure}[b]{0.21\textwidth}
         \centering
         \includegraphics[width=\textwidth]{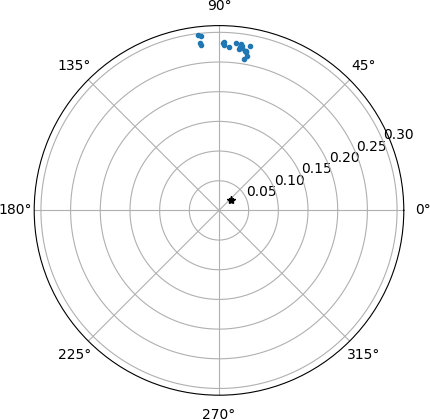}
         \caption{Non-IID, $t = 64$}
     \end{subfigure}
    \caption{\footnotesize The distribution of local updates for AlexNet on IID and Non-IID data at different communication rounds for Cifar-10 dataset. Each blue dot corresponds to the local update from one client. The black dot shows the magnitude and the cosine angle of global local model update at iteration $t$.}
    \label{fig:app_alex_cifar}
\end{figure}

\begin{figure}[t!]
    \centering
     \begin{subfigure}[b]{0.21\textwidth}
         \centering         
		 \includegraphics[width=\textwidth]{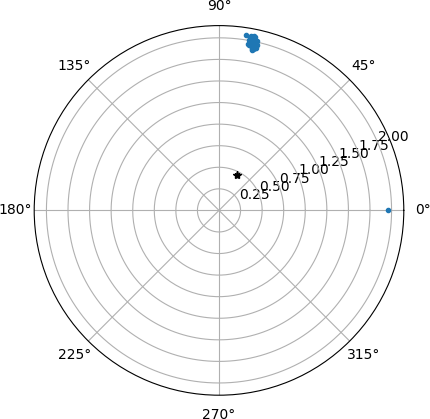}
         \caption{IID, $t = 0$}
     \end{subfigure}
     \hfill
     \begin{subfigure}[b]{0.21\textwidth}
         \centering
         \includegraphics[width=\textwidth]{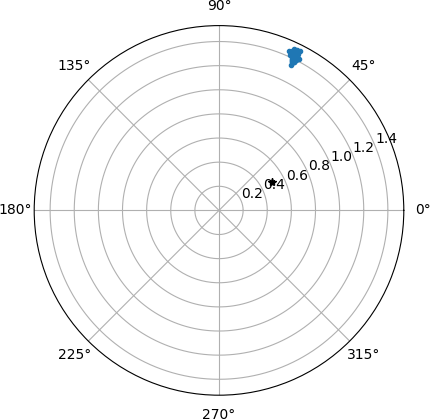}
         \caption{IID, $t = 2$}
     \end{subfigure}
     \begin{subfigure}[b]{0.21\textwidth}
         \centering
         \includegraphics[width=\textwidth]{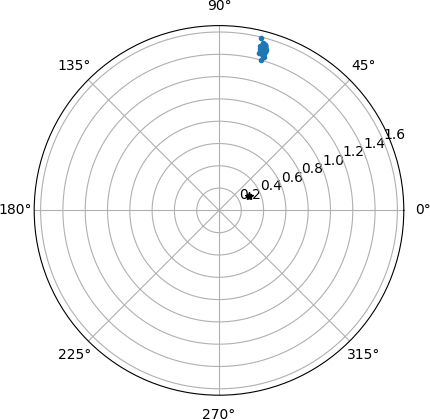}
         \caption{IID, $t = 8$}
     \end{subfigure}
     \hfill
     \begin{subfigure}[b]{0.21\textwidth}
         \centering
         \includegraphics[width=\textwidth]{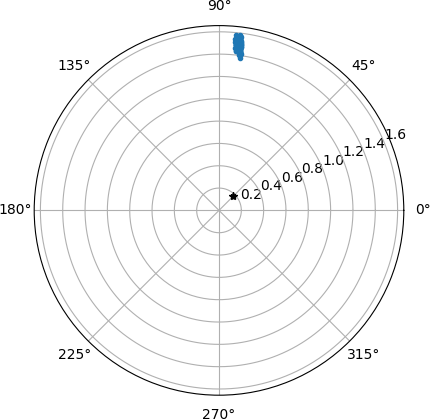}
         \caption{IID, $t = 32$}
     \end{subfigure}
     \begin{subfigure}[b]{0.21\textwidth}
         \centering         
		 \includegraphics[width=\textwidth]{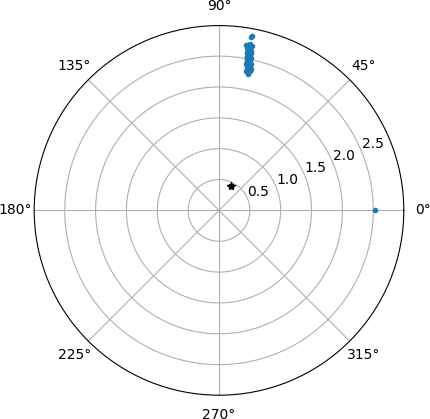}
         \caption{Non-IID, $t = 0$}
     \end{subfigure}
     \hfill
     \begin{subfigure}[b]{0.21\textwidth}
         \centering
         \includegraphics[width=\textwidth]{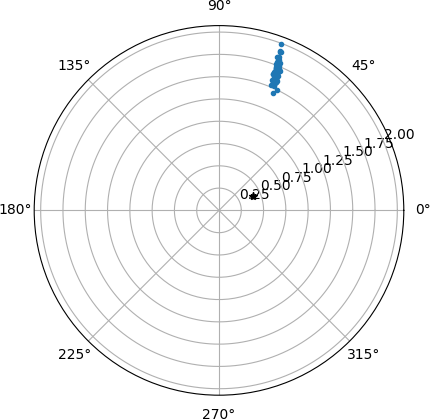}
         \caption{Non-IID, $t = 2$}
     \end{subfigure}
     \begin{subfigure}[b]{0.21\textwidth}
         \centering
         \includegraphics[width=\textwidth]{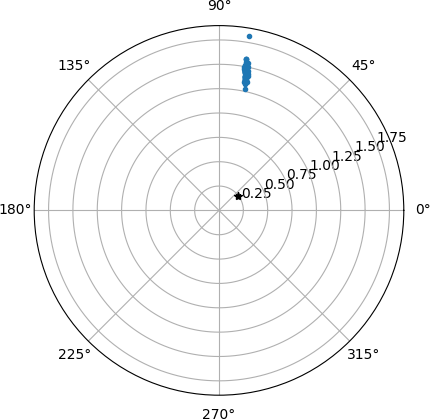}
         \caption{Non-IID, $t = 8$}
     \end{subfigure}
     \hfill
     \begin{subfigure}[b]{0.21\textwidth}
         \centering
         \includegraphics[width=\textwidth]{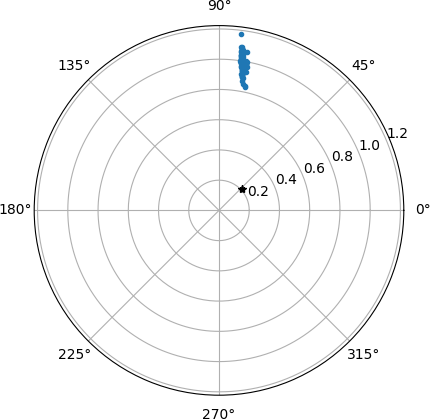}
         \caption{Non-IID, $t = 32$}
     \end{subfigure}
    \caption{\footnotesize The distribution of local updates for ResNet-18 on IID and Non-IID data at different communication rounds for Cifar-10 dataset. Each blue dot corresponds to the local update from one client. The black dot shows the magnitude and the cosine angle of global local model update at iteration $t$.}
    \label{fig:app_resnet_cifar}
\end{figure}

In this part, we plot the change of the distributions of the update differences of different algorithms listed in the main paper.  Notice that in all models and datasets, the distributions of the magnitude in the IID cases are more concentrated than the corresponding Non-IID cases. Also, the distributions of the same model trained on EMNIST dataset are more concentrated than trained on Cifar-10 dataset.
\end{document}